\documentclass[final,onefignum,onetabnum]{article}

\usepackage{array}

\usepackage[caption=false]{subfig}
\usepackage{pgfplots}

\usepackage{algorithmic}

\usepackage{graphicx,epstopdf}

\usepackage{subfig}
\usepackage{amsmath}
\usepackage{amssymb}
\usepackage{graphicx}
\usepackage{dcolumn}
\usepackage{mathtools}
\usepackage{amsmath,amsfonts}
\usepackage{bm}
\usepackage{amsmath}
\usepackage{amssymb}
\usepackage{color}
\usepackage{float}
\usepackage{setspace}
\usepackage{tabularx}

\allowdisplaybreaks

\newcommand{\E}{ {\mathbb{E}} }

\newcommand{\be}{\begin{equation}}
\newcommand{\ee}{\end{equation}}

\def\ba{\begin{array}}                \def\ea{\end{array}}
\def\bel{\begin{equation}\label}      \def\ee{\end{equation}}

\colorlet{texcscolor}{blue!50!black}
\colorlet{texemcolor}{red!70!black}
\colorlet{texpreamble}{red!70!black}
\colorlet{codebackground}{black!25!white!25}

\date{}

\title{A Backward SDE Method for Uncertainty Quantification in Deep Learning}

\author{
Richard Archibald\thanks{ Computational Science and Mathematics Division, Oak Ridge National Laboratory, Oak Ridge, Tennessee.}
\and Feng Bao\thanks{ Department of Mathematics, Florida State University, Tallahassee, Florida, \ ({\tt bao@math.fsu.edu}).}
\and Yanzhao Cao \thanks{ Department of Mathematics, Auburn University, Auburn, Alabaama}
\and He Zhang \thanks{ School of Mathematics, Jilin University, Changchun, China, }
      }

\begin{document}
\maketitle

\begin{abstract}
 We develop a probabilistic machine learning method, which formulates a class of stochastic neural networks by a stochastic optimal control problem. An efficient stochastic gradient descent algorithm is introduced under the stochastic maximum principle framework. Numerical experiments for applications of stochastic neural networks are carried out to validate the effectiveness of our methodology.  
\end{abstract}


\textbf{AMS:} 60H35, 68T07, 93E20	

\section{Introduction}

In this paper, we introduce an efficient computational framework to quantify the uncertainty of a class of deep neural networks (DNNs). The DNN has emerged from machine learning and becomes one of the most extensively studied research topics in scientific computing and data science. As a type of artificial neural network with multiple layers, the DNN is capable to model complex systems and its applications cover wide range among various scientific and engineering disciplines.
However, despite phenomenal success, the deterministic output of DNNs can not produce probabilistic predictions for the uncertain nature of scientific research.
Therefore, it's difficult to apply DNNs to solve real-world scientific problems. 
To address the challenge of uncertainty quantification, probabilistic machine learning approaches are developed based on probabilistic modeling \cite{Nature-Learning}. The state-of-the-art probabilistic learning approach is the Bayesian neural network (BNN) \cite{BNN-ICML-2015, BNN-NIPS2016}. The main idea of the BNN is to treat parameters in a DNN as random variables, and then approximate distributions of these random parameters.  Instead of searching for the optimal parameters by deterministic optimization, the BNN utilizes Bayesian optimization to derive parameter distributions. As a result, the estimated random parameters generate random output, which characterize the uncertainty of the target model.
Although the BNN approach provides a neat conceptual framework to quantify the uncertainty in probabilistic machine learning, carrying out Bayesian optimization to estimate a large number of parameters in the BNN poses several computational challenges such like high dimensional integration, high dimensional optimization, and high dimensional approximation. 
In recent studies, the Markov Chain Monte Carlo (MCMC) sampling method is introduced to address the challenge of high dimensional integration in Bayesian inference, and a class of gradient based methods, such like Langevin MCMC \cite{Langevin-MC} and Hamiltonian MCMC \cite{Hamiltonian-MC}, are applied to improve the efficiency of high dimensional optimization. However, the primary computational challenge of approximating high dimensional distributions still remains. Thus training BNNs is still a computationally extensive task.

In this work, we focus on an alternative probabilistic learning approach, which allows us to avoid Bayesian optimization. 
The neural network architecture in our approach is an extension of the so-called ``Neural ODE'', which formulates the evolution of hidden layers in the DNN as a discretized ordinary differential equation (ODE) system \cite{chen2018neural}. The Neural ODE formulation of deep learning provides a mathematical description for the residual neural network, which is an important neural network structure in machine learning. 
To incorporate uncertainties in the neural network, instead of treating parameters as random variables in the Bayesian approach, we add Gaussian-type noises to the hidden layers and construct a stochastic ordinary differential equation (SDE) formulation of DNN, which will be called the stochastic neural network (SNN) in this paper. Such a stochastic construction of neural network has been extensively studied recently \cite{Neural-Jump-SDE, SDE-Net, Neural-SDE}. 
The major difference between the SNN and the BNN is that the uncertainty of SNN is generated by the diffusion term of SDE and the coefficient of diffusion controls the probabilistic output of SNN. Therefore, the point estimation for diffusion coefficients in the SNN approach could  achieve the goal of uncertainty quantification for probabilistic learning.
While the construction of SDE type neural networks and justification for SNNs are well accepted, the training process is also challenging for SNNs. In the deterministic DNN, the optimal parameters are determined by (stochastic) gradient descent optimization, and the gradient is calculated by the chain rule. 
However, when trying to find the gradient with respect to parameters in an SDE, the standard chain rule is not applicable due to the stochastic integrals, and It\^o calculus is needed, which makes the derivation of the gradient complicated. To derive a mathematical expression for the gradient, we formulate the SNN model as a \textit{stochastic optimal control problem}. Specifically, we consider the state of neurons in SNN as controlled state, and the SNN parameters are considered as control terms that govern the controlled state to meet the optimal cost condition, which is chosen to be the minimum loss for the SNN output. In this way, the training process for the SNN, which seeks the optimal parameters, is equivalent to the solving process for the stochastic optimal control problem, which determines the optimal control.

The theoretical framework that we adopt to solve the SNN version of stochastic optimal control problem is the stochastic maximum principle (SMP) \cite{Peng_control} due to its advantage in solving high dimensional problems --- compared with its alternative approach, i.e. the dynamic programming principle \cite{Yong-Zhou-1999, Control_survey}. Therefore, our method can also be considered as an \textit{``SMP approach for SNNs''}. The central idea of SMP is that a stochastic optimal control problem must satisfy an optimality condition of a function called the Hamiltonian, which consists of solutions of an adjoint backward SDE (BSDE). In this way, the gradient with respect to the optimal control is expressed by solutions of the adjoint BSDE, and hence solving the stochastic optimal control problem through SMP requires obtaining solutions of the BSDE at each gradient descent iteration step. Although several successful numerical methods are developed to solve BSDEs \cite{Four_step, ZhangJ_BSDE, Milstein_BSDE, Zhao_BSDE, ML_BSDE}, approximating solutions of BSDEs in the high dimensional (controlled) state space at each iteration step is still extremely challenging.

The main theme of our computational method to implement the SMP approach for SNNs is to solve for the optimal control, which is equivalent to the optimal parameters in SNNs, under the SMP framework without solving the adjoint BSDE in the high dimensional state space.
The key to the success of our method is the fact that solutions of the adjoint BSDE are used to formulate the gradient, which will be used in the optimization procedure for the optimal control, and obtaining solutions of BSDEs in the entire state space is not the goal of optimization. In other words, the adjoint BSDE provides a mechanism for the ``backpropogation'' in the training process of SNN, and the optimal parameters of the SNN are determined by gradient-based optimization, where the gradient consists of solutions of the adjoint BSDE.
In data driven optimization problems, gradient-based optimization is often carried out by stochastic gradient descent (SGD), which utilizes one data sample to represent the entire data set.
A novel concept that we shall introduce in this work is that the random samples in the state space that characterize solutions of the adjoint BSDE can be considered as ``pseudo-data'' in the optimization procedure, and we extend the application of SGD by treating the state samples of the adjoint BSDE as a source of data. In this connection, we apply SGD to both the real training data and the pseudo state sample data in the optimization procedure, and we select one training data and one sample data each time to represent their corresponding data sets.
As a result, at each SGD iteration step, we only pick one sample-path in the state space and solve the BSDE along the chosen sample-path. In this way,  we avoid solving the BSDE in the entire high dimensional state space repeatedly at each gradient descent iteration step, which makes our SGD optimization an efficient numerical method to implement the SMP approach for SNNs.

\vspace{0.2em}

The rest of this paper is organized as following. In Section \ref{SMP for SNN}, we introduce the mathematical foundation of our SMP approach for SNNs. In Section \ref{SGD-SMP}, we derive an SGD optimization algorithm to solve the stochastic optimal control problem under the SMP framework, which is equivalent to the training algorithm that determines the desired optimal parameters in the SNN. Numerical experiments will be presented in Section \ref{Numerics} to examine the performance of our algorithm in three applications of SNNs, and we will give some concluding remarks and the plan of future work in Section \ref{Conclusion}.

%
%
%
%
%
%
%


\section{Stochastic maximum principle approach for stochastic neural networks}\label{SMP for SNN}
In this section, we introduce the mathematical framework of our stochastic maximum principle approach for stochastic neural networks (SNNs). The main theme of our approach is to formulate a class of SNNs by stochastic differential equations, which describe stochastic forward propagation of deep neural networks (DNNs), and then treat the training process as a stochastic optimal control problem, which will be solved by stochastic maximum principle. 

\subsection{Stochastic neural networks}

The SNN structure that we consider in this work is given by the following model
\begin{equation}\label{Intro:SNN}
X_{n+1} = X_n + h F(X_n, \theta_n ) + \sqrt{h} \sigma_n \omega_n, \qquad n = 0, 1, 2, \cdots, N-1,
\end{equation}
where $X_n : = [x^1, x^2, \cdots x^L] \in \mathbb{R}^{L \times d}$ denotes the vector that contains all the $L$ neurons at the $n$-th layer in a DNN; $F$ is the pre-chosen activation function, which is commonly picked among the sigmoid function, the hyperbolic tangent function, and the rectified linear unit function (ReLu); $h$ is a fixed positive constant that stabilizes the network;  $\theta_n$ represents neural network parameters (such like weights and biases), which determine the output of the neural network and will be learned through the ``training process''; $\{\omega_n\}_n : = \{\omega_n\}_{n=0}^{N-1}$ is a sequence of i.i.d. standard Gaussian random variables, together with the coefficient matrices $\{\sigma_n\}_n$, the stochastic terms $\{\sigma_n \omega_n\}_n$ generate artificial noises that bring uncertainties to a deterministic DNN, which allows the SNN to produce random output to reflect stochastic behaviors of the target model. The initial state $X_0 \in \mathbb{R}^d$ in \eqref{Intro:SNN} is the input variable in a neural network and $X_N$ is the output. Different from the Bayesian neural network (BNN) formulation of probabilistic machine learning, which let the parameter $\theta$ in a standard DNN be a random variable and apply Bayesian inference to generate the empirical distribution of the parameter from training data, we introduce uncertainties to the SNN model by adding a random noise term. 

When $\sigma_n = 0$, the above SNN model reduces to a standard DNN, which can be formulated by the so-called ``neural ordinary differential equation''  (Neural ODE). 
The main concept of Neural ODE is to describe a DNN as a discretized ODE system, where standard analysis for ODEs can be applied, and the training process for a DNN is equivalent to solving a deterministic optimal control problem \cite{weinan2019mean}. Following the argument in Neural ODE, we consider the SNN model in the form of the stochastic sequence \eqref{Intro:SNN} as a discretized stochastic (ordinary) differential equation (SDE), where the activation function $F$ defines the drift coefficient and $\sigma$ is the diffusion coefficient, and the propagation of the SDE system implements the forward propagation in the SNN. In connection with the optimal control perspective on DNN, the training process for SNN is equivalent to a stochastic optimal control problem.
In what follows, we introduce our stochastic optimal control formulation of SNN. 


\subsection{Stochastic optimal control formulation of stochastic neural networks}
We consider the following continuous form of the SNN model \eqref{Intro:SNN} in a complete filtered probability space $(\Omega, \mathcal{F}, \mathbb{F}^W, \mathbb{P})$
\begin{equation}\label{SNN-SDE}
X_{t} = X_0 + \int_{0}^{T} F(X_s, \theta_s )ds + \int_0^T \sigma_s dW_s,
\end{equation}
where $W:= \{W_t\}_{0 \leq t \leq T}$ is a standard Brownian motion (corresponding to the Gaussian noise $\{\omega_n\}_n$) and $\sigma$ is the diffusion coefficient. The stochastic integral $\int_0^T \sigma_s dW_s$ on the right hand side of the equation \eqref{SNN-SDE} is an It\^o type integral, which brings uncertainties to the forward propagation in the SNN. The temporal index $T>0$ is a given positive constant that represents the output layer of the neural network. When a specific temporal discretization on the interval $[0, T]$ is chosen, the depth of the SNN is determined.
In this work, instead of treating $\theta$ in the activation function $F$ as a parameter to be optimized in the learning process, we consider $\theta$ and $\sigma$ as control terms in a stochastic optimal control problem. Since the uncertainties in probabilistic learning and stochastic behaviors of the SNN model are described by the stochastic integral in \eqref{SNN-SDE}, we let both $\theta$ and $\sigma$ be deterministic control processes in this work for efficiency of optimization.
For convenience of presentation,  we let $u := [\theta, \sigma]$ be the control vector in an admissible control set denoted by $\mathcal{U}[0, T]$, and we rewrite the continuous SNN model \eqref{SNN-SDE} as the following controlled process in its differential form
\begin{equation}\label{Controlled-State}
dX_{t} = f(X_t, u_t )dt + g(u_t) dW_t,  \qquad 0 \leq t \leq T,
\end{equation}
where $f(X_t, u_t) = F(X_t, \theta_t)$, $g(u_t) = \sigma_t$, 
and $X_t$ is usually called the \textit{``state process''}. When the stochastic optimal control framework is used to formulate a machine learning problem, the control $u$ aims to minimize the discrepancy between the SNN output and the data. Therefore, we define the cost function in our stochastic optimal control problem as
\begin{equation}\label{Control-Cost}
J(u) := \E[\Phi(X_T, \Gamma)],
\end{equation}
where $\Gamma$ is a random variable that generates training data in machine learning, which also depends on $X_0$, and $\Phi(X_T, \Gamma): = \|X_T - \Gamma\|_{Loss}$ is a loss function with its corresponding error norm $\| \cdot \|_{Loss}$ \cite{Haber_2017}. 
The goal of stochastic optimal control problems is to find an \textit{optimal control} $\bar{u}$ that minimizes the cost $J$, i.e. find $\bar{u}$ such that
\begin{equation}\label{Optimal-Control}
J(\bar{u}) = \inf_{u \in \mathcal{U}[0, T]} J(u).
\end{equation}
In this way, the optimal control $\bar{u}$ also minimizes the difference between $X_T$ and $\Gamma$ in the loss function, and hence $\bar{u}$ is equivalent to the desired optimal parameters in the SNN.

There are two well-known approaches to solve the stochastic optimal control problem -- the dynamic programing and the stochastic maximum principle. The main theme of the dynamic programming approach is to solve the stochastic optimal control problem through numerical solutions for the Hamilton-Jacobi-Bellman (HJB) equation, which is a nonlinear partial differential equation (PDE) \cite{Feng_2013}. On the other hand, the stochastic maximum principle approach aims to seek the optimal control that satisfies an optimality condition of a function called the (stochastic) Hamiltonian, and it's typically achieved by gradient descent type optimization methods \cite{Gong_2017}. In most machine learning applications, neural networks contain large number of neurons, and the number of neurons corresponds to the dimension of HJB equations. Therefore, it is very difficult to apply the dynamic programming approach to solve stochastic optimal control problems that we use to formulate the SNN model --- due to the ``curse of dimensionality'' of solving high dimensional PDEs, and we adopt the stochastic maximum principle approach in this work.


\subsection{Stochastic maximum principle approach for the stochastic optimal control}

In stochastic maximum principle (SMP), we assume that the optimal control $\bar{u}$ is in the interior of $\mathcal{U}[0, T]$. Then, by applying the G\^ateaux derivative of $J$ with respect to $u$, one can derive that the gradient process of the control function $J$ with respect to the control process on the interval $[0, T]$ has the following expression (see \cite{Yong-Zhou-1999})
\begin{equation}\label{Intro:Gradient}
J'_u(t, u_t) = \E\big[f'_u(X_t, u_t)^T Y_t + g_u'(u_t)^T Z_t\big],
\end{equation}
where $X$ is the controlled state process introduced in \eqref{Controlled-State}, $Y$ and $Z$ satisfy the following \textit{adjoint} backward SDE, which is corresponding to the (forward) state process $X_t$,
\begin{equation}\label{Intro:BSDE}
dY_t = - f'_x(X_t, \bar{u}_t)^T Y_t dt + Z_t dW_t, \qquad Y_T = \Phi'_x(X_T, \ \Gamma).
\end{equation}
The pair $(Y, Z)$ is called the adapted solutions of the above adjoint backward SDE (BSDE), where $Y$ propagates \textit{backwards} from $T$ to $0$ with initial condition $Y_T = \Phi'_x(X_T, \ \Gamma)$, and $Z$ is the martingale representation of $Y$ with respect to $W$. An important property of the BSDE \eqref{Intro:BSDE} for the continuous SNN model \eqref{Controlled-State} is that the solutions $Y_t$ and $Z_t$ are functions of $X_t$ given the state $X_t$ \cite{BSDE_finance}.

With the gradient process $J'_u$, the standard approach to find the optimal control $\bar{u}$ is the gradient descent optimization. Specifically, for a pre-chosen initial guess $u^0$, we carry out the following gradient descent iteration
\begin{equation}\label{u-GD}
u_t^{k+1} = u_t^k - \eta_k J'_u(t, u_t^k), \qquad k = 0, 1, 2, \ \cdots, \quad 0 \leq t \leq T,
\end{equation}
where $\eta_k>0$ is the step-size of the gradient.
Since the stochastic optimal control problem \eqref{Controlled-State} - \eqref{Optimal-Control} is designed to formulate the SNN model, the optimization procedure to find the optimal control $\bar{u}$ is equivalent to the training process to determine the ``optimal parameter'' that fits the training data, and the gradient descent step-size $\eta_k$ is also the ``learning rate'' in machine learning. 
We want to mention that in supervised learning, there's an essential procedure called ``backpropagation'', which computes the gradient of the loss function with respect to network parameters. However, due to the stochastic nature of SNN, mathematical tools (such like the chain rule) that derive the gradient in the deterministic backpropagation are not applicable. 
In this way, the BSDE \eqref{Intro:BSDE} in the SMP approach can be considered as a stochastic version of backpropagation in machine learning that generates the gradient process.

\section{An efficient stochastic gradient descent algorithm for stochastic maximum principle}\label{SGD-SMP}
Our numerical schemes are constructed on discrete points over the interval $[0, T]$ with a discretization defined by
$$
\Pi^N : = \Big\{t_n | 0=t_0 < t_1 < t_2 < \cdots t_n < t_{n+1} < \cdots < t_N=T \Big\},
$$
where $N$ is the partition number, which is equivalent to the depth of SNNs. When $\Pi^N$ is a uniform partition, we have $h = \frac{T}{N}$. In most DNN models, the step-size $h$ is chosen to be equal to $1$ --- although some analysis outcomes in Neural ODE show that appropriately designed step-sizes may stabilize DNNs \cite{Haber_2017, CNN_Control}. 
To develop our stochastic gradient descent algorithm for searching the optimal control, we shall first provide numerical solutions for BSDEs.

\subsection{Numerical solutions for backward stochastic differential equations}\label{Numerical_BSDE}
Our numerical schemes for BSDEs consist of numerical simulation for the controlled state process \eqref{Controlled-State} and numerical solutions for the adjoint BSDE \eqref{Intro:BSDE}. The derivation of numerical schemes is based on approximation for integrals on sub-intervals $[t_n, t_{n+1}]$, $0 \leq n \leq N-1$.

Since the controlled state process is a standard forward SDE (with the control term $u$), we apply the Euler-Maruyama scheme and derive the following approximation equation for $X$ on the sub-interval $[t_n, t_{n+1}]$,
\begin{equation}\label{discrete:dX}
X_{t_{n+1}} = X_{t_n} + h  f(X_{t_n}, u_{t_n}) + g(u_{t_n}) \Delta W_{t_n} + R_X^n,  \qquad 0 \leq n \leq N-1,
\end{equation}
where $\Delta W_{t_n}:= W_{t_{n+1}} - W_{t_n}$ and 
$R_X^n : = \int_{t_n}^{t_{n+1}} f(X_s, u_s) ds -  h f(X_{t_n}, u_{t_n})  + \int_{t_n}^{t_{n+1}} g(u_s) d W_s -  g(u_{t_n}) \Delta W_{t_n}$ is the approximation error term. 
We can see that the approximation scheme \eqref{discrete:dX} (without the approximation error term $R_X^n$) is equivalent to the discrete SNN model \eqref{Intro:SNN} given the definitions of $f$ and $g$ in the continuous controlled process \eqref{Controlled-State}, and the control term $u_{t_n}$ plays the role of parameters $\theta_n$ and $\sigma_n$. Therefore, the forward simulation of the state process $X$ in our stochastic optimal control problem coincides the forward propagation of the SNN model.

To derive numerical schemes for the adjoint BSDE, we consider the equation \eqref{Intro:BSDE} on the sub-interval $[t_n, t_{n+1}]$, $0 \leq n \leq N-1$, i.e.
\begin{equation}\label{discrete:BSDE}
Y_{t_n} = Y_{t_{n+1}} + \int_{t_n}^{t_{n+1}} f_x'(X_s, u_s) Y_s ds - \int_{t_n}^{t_{n+1}} Z_s dW_s.
\end{equation}
We take the conditional expectation $\E_{n}^X[\cdot] : = \E[\cdot | X_{t_n}]$ on both sides of the above equation and approximate the deterministic integral by using the right-point formula to get
\begin{equation}\label{Exp:Y}
Y_{t_n} = \E_{n}^X\big[Y_{t_{n+1}} \big] + h \E_{n}^X\big[ f_x'(X_{t_{n+1}}, u_{t_{n+1}}) Y_{t_{n+1}} \big] + R_Y^n,  
\end{equation}
where we have used the fact  $ Y_{t_n}  = \E_{n}^X\big[ Y_{t_n}\big]$ due to the adaptedness of the solution $Y$ with respect to $W$, and the stochastic integral becomes $0$ under the expectation $\E_{n}^X[\cdot]$, i.e.
$\E_{n}^X[\int_{t_n}^{t_{n+1}} Z_s dW_s] = 0$.
The error term 
$R_Y^n : = \E_n^X\Big[ \int_{t_n}^{t_{n+1}} f_x'(X_s, u_s) ds -   f_x'(X_{t_{n+1}}, u_{t_{n+1}}) Y_{t_{n+1}}h \Big]$ 
on the right hand side of \eqref{Exp:Y} contains the approximation error for the deterministic integral, 
and we can derive a numerical scheme to solve for $Y_{t_n}$ by dropping the error term $R_Y^n$.

From the above discussion, we can see that the stochastic integral term contains the solution $Z$ and it's eliminated by the expectation. To maintain the stochastic integral and get a numerical scheme for $Z$, we use the left-point formula to approximate both the deterministic integral and the stochastic integral in \eqref{discrete:BSDE}. As a result, we obtain
\begin{equation}\label{discrete:BSDE_dW}
Y_{t_n} = Y_{t_{n+1}} + h f_x'(X_{t_n}, u_{t_n}) Y_{t_n} - Z_{t_{n}} \Delta W_{t_n} + R_{Z}^n,
\end{equation}
where 
$R_{Z}^n =  \int_{t_n}^{t_{n+1}} f_x'(X_s, u_s) ds -   f_x'(X_{t_{n}}, u_{t_{n}}) Y_{t_{n}}h - \int_{t_n}^{t_{n+1}} Z_s dW_s + Z_{t_{n}} \Delta W_{t_n}$
is the approximation error term. Then we multiply $\Delta W_{t_n}$ on both sides of \eqref{discrete:BSDE_dW} and take conditional expectation $\E_{n}^X[\cdot]$. 
Since $X$ and $Y$ are adapted to $W$, we know that $\E_n^X[ Y_{t_{n}} \Delta W_{t_n}] = 0$ and $\E_n^X[ f_x'(X_{t_{n}}, u_{t_{n}}) Y_{t_{n}}h \Delta W_{t_n}] = 0$. 
Therefore, the equation \eqref{discrete:BSDE_dW} becomes
\begin{equation}\label{Exp:Z}
\E_n^X\big[ Z_{t_{n}}\big] h = \E_n^X[ Y_{t_{n+1}} \Delta W_{t_n}] + \E_n^X[R_{Z}^n \Delta W_{t_n}],
\end{equation}
where the left hand side of the above equation is obtained by the fact that $Z$ is also $W$ adapted, which gives us 
$$\E_n^X\big[ Z_{t_{n}} (\Delta W_{t_n})^2\big] =  \E_n^X\big[ Z_{t_{n}}\big] h. $$

By dropping the approximation error terms $R_x^n$ and $R_{Y}^n$ in \eqref{discrete:dX} and \eqref{Exp:Y}, respectively, and dropping the error $\E_n^X[R_{Z}^n \Delta W_{t_n}]$ in \eqref{Exp:Z}, we obtain our numerical method to solve the BSDE \eqref{Intro:BSDE}: 
\vspace{0.75em}

\noindent For $n = N-1, N-2, \cdots, 2, 1, 0$, we solve the BSDE \eqref{Intro:BSDE} with the following schemes
\begin{equation}\label{Semi-BSDE}
Y_n = \E_n^X[ Y_{n+1}] + h \E_n^X\big[ f_x'(X_{n+1}, u_{t_{n+1}}) Y_{n+1} \big], \qquad
Z_n = \frac{\E_n^X[ Y_{n+1} \Delta W_{t_n}]}{h}, 
\end{equation}
where $X_{n+1}$ is the approximation for the forward controlled state $X_{t_{n+1}}$, which is obtained by the following discrete scheme for the state process, i.e.
\begin{equation}\label{Semi-SDE}
X_{n+1} = X_{n} + f(X_n, u_{t_n}) h + g(u_{t_n}) \Delta W_{t_n}, 
\end{equation}
and $Y_n$, $Z_n$ are numerical approximations for $Y_{t_n}$, $Z_{t_n}$, respectively. In practical simulations, we let $\Delta W_{t_n} = \sqrt{h} \omega_n$, where $\omega_n$ is the Gaussian random variable in the SNN model \eqref{Intro:SNN}.

\vspace{0.75em}
In our SMP approach for SNNs, the scheme \eqref{Semi-SDE} has the same formulation of the $N$-layer SNN model, and the simulation of $X_n$ from $n = 0$ to $n = N$ is equivalent to the forward propagation of SNNs, where the input layer of the SNN is $X_0$ and the output layer of the SNN is $X_N$.
At the same time, the numerical schemes \eqref{Semi-BSDE} for the adjoint BSDE provide a mechanism for the backpropagation of SNNs in the training process since they compose the gradient with respect to the control, which is equivalent to SNN parameters.  In this way, numerical schemes \eqref{Semi-BSDE}-\eqref{Semi-SDE} give us a computational framework to train the SNN model introduced in \eqref{Intro:SNN}. 
\vspace{0.5em}

With numerical solutions $\{ Y_n \}_n$ and $\{ Z_n\}_n$ obtained in the schemes \eqref{Semi-BSDE}, we have the approximated gradient with respect to control as following
\begin{equation}\label{Semi_Approx-J'}
\bar{J}'_u(n, u_{t_n}) = \E\big[f'_u(X_{n}, u_{t_n})^T Y_n + g_u'(u_{t_n})^T Z_n\big], \qquad n = 0, 1, 2, \cdots, N-1,
\end{equation}
where $\bar{J}'_u(n, u_{t_n})$ is an approximation for $J'_u(t_n, u_{t_n})$ by using approximated solutions $X_n$, $Y_n$ and $Z_n$. Then, the gradient descent iteration on the discretization $\Pi^N$ can be carried out as following
\begin{equation}\label{Semi_Approx-GD}
\begin{aligned}
u_n^{k+1} = & u_n^k - \eta_k \bar{J}'_u(n, u_n^k) \\
= & u_n^k - \eta_k \E\big[f'_u(X^k_{n}, u_{n}^k)^T Y^k_n + g_u'(u_{n}^k)^T Z^k_n\big], \quad n = 0,1, \cdots, N-1
\end{aligned}
\end{equation}
where $X_n^k$, $Y_n^k$ and $Z_n^k$ are obtained under the control $\{u_n^k\}_n$, i.e. choose $u_{{t_n+1}} = u_{n+1}^k$ and $u_{{t_n}} = u_{n}^k$ in schemes \eqref{Semi-BSDE} and \eqref{Semi-SDE} respectively. Then, for a pre-chosen integer $K$ as our stopping criteria, we let $u_n^K$ be our estimate for the optimal control $\bar{u}_{t_n}$.

\vspace{0.3em}
In order to implement the schemes \eqref{Semi-BSDE}-\eqref{Semi_Approx-GD}, one needs to approximate the (conditional) expectations. Well-known numerical methods for approximating expectations include numerical integrations and Monte Carlo simulations.
 Since our computational framework for stochastic optimal control problems is introduced to solve the SNN problem, the controlled state $X$, which is used to model stacked neurons, is typically a high dimensional random variable. Therefore,     numerical integration methods are usually not feasible. In what follows, we introduce the Monte Carlo method to approximate the expectations as a fully calculated approach for the gradient descent iteration \eqref{Semi_Approx-GD}. 
To proceed, we first use the scheme \eqref{Semi-SDE} to generate simulations of $X$ as Monte Carlo samples corresponding to the controlled state process, i.e.
\begin{equation}\label{Full-SDE}
X_{n+1}^{m, k} = X_{n}^{m, k} + f(X_n^{m, k}, u_{n}^k) h + g(u_{n}^k) \sqrt{h} \omega_n^m, \qquad 0 \leq n \leq N-1, \quad m = 1, 2, \cdots, M,
\end{equation}
where $M$ is the number of Monte Carlo samples that we use to describe the state $X$, and $\omega_n^m$ represents the $m$-th realization of the Gaussian random variable $\omega_n$. 
Then, we approximate conditional expectations in \eqref{Semi-BSDE} by Monte Carlo averages, i.e.
\begin{equation*}\label{MC-YZ}
\begin{aligned}
\E^X_n[Y_{n+1}^k] \approx \sum_{m=1}^M\frac{ Y^k_{n+1}(X_{n+1}^{m, k})}{M}, \quad &
\E^X_n[f_x'(X_{n+1}, u_{n+1}^k) Y_{n+1}^k] \approx  \sum_{m=1}^M\frac{ f_x'(X_{n+1}^{m, k}, u_{n+1}^k) Y^k_{n+1}(X_{n+1}^{m, k})}{M}, \\
\E^X_n[Y_{n+1}^k \Delta W_{t_n}] \approx &  \sum_{m=1}^M\frac{ Y^k_{n+1}(X_{n+1}^{m, k}) \sqrt{h} \omega_n^m }{M}.
 \end{aligned}
\end{equation*}
It's worthy to point out that the side condition of the adjoint BSDE \eqref{Intro:BSDE}, i.e. $Y_T = \Phi'_x(X_T, \Gamma)$, contains the random variable $\Gamma$, which describes data. Hence the numerical schemes that we introduced to solve the adjoint BSDE need to be carried out on each data sample $\gamma \sim \Gamma$, and an extra Monte Carlo average is needed for the data variable $\Gamma$ when calculating the expectation in the gradient descent iteration \eqref{Semi_Approx-GD}.
As a result, we use two layers of Monte Carlo simulation (for the state process $X$ and for the data variable $\Gamma$)
to approximate the expectation and rewrite the gradient descent iteration as
\begin{equation}\label{Full_Approx-GD}
u_n^{k+1} = u_n^k - \eta_k \sum_{q=1}^{Q}\sum_{m=1}^M \frac{f'_u(X_{n}^{m, k}, u_{n}^k)^T Y^{k}_n|_{\gamma_q}(X_{n}^{m, k}) + g_u'(u_{n}^k)^T Z^k_n|_{\gamma_q}(X_{n}^{m, k})}{M Q},
\end{equation}
where $Y^{k}_n|_{\gamma_q}$ and $Z^{k}_n|_{\gamma_q}$ denote approximated solutions $Y^k_n$ and $Z^k_n$ corresponding to the data sample $\gamma_q \sim \Gamma$ in the side condition, and $Q$ is the total number of  available training data, which can be considered as Monte Carlo samples that describe the data variable $\Gamma$.

\vspace{0.5em}

Although the Monte Carlo approach is the state-of-the-art method to approximate expectations, when the dimension of the controlled state $X$ is high, which is common in SNN applications, and when the discretization number $N$ is large, i.e. the SNN is deep, the Monte Carlo sampling number $M$ needs to be very large, which makes the computational cost to implement the gradient descent optimization \eqref{Full_Approx-GD} very high.
Moreover, it's important to recall that the values of $Y$ and $Z$ correspond to the state of $X$. Therefore we also need to derive mappings from $X$ to solutions $Y$ and $Z$, which can be considered as spatial approximation for solutions of BSDEs. In most existing numerical methods, this is accomplished by function approximation. Specifically, a set of spatial points that describe the random variable $X$ are chosen, and people typically use approximated solution values on those pre-chosen spatial points to construct interpolatory approximations for solutions in the state space. In this way, the function values $Y_{n}^k|_{\gamma_q}(X_{n}^{m, k})$ and $Z_{n}^k|_{\gamma_q}(X_{n}^{m, k})$ in the scheme \eqref{Full_Approx-GD} are calculated by numerical interpolation methods, such like polynomial interpolations, sparse grid interpolations, or meshfree approximations. Recently, several machine learning type approximation methods are developed, which give us effective global  meshfree approximators to simulate functions \cite{ML-Approximation18, Learning-approximation}. In spite of the fact that function approximation is a well-established field and conventional numerical methods are very successful, obtaining a complete approximation for a high dimensional function is still an extremely challenging task. Especially, in the gradient descent optimization procedure \eqref{Full_Approx-GD}, we need to solve the BSDE system \eqref{Intro:BSDE} for every data $\gamma_q$ that we pick as the side condition $Y_T= \Phi'_x(X_T, \gamma_q)$ to get one update for the estimated optimal control. Thus complete numerical implementations for the iteration scheme \eqref{Full_Approx-GD} is computationally expensive.

To address the aforementioned computational challenges in Monte Carlo simulation and high dimensional function approximation, in the next subsection, we introduce an efficient stochastic gradient descent algorithm to carry out the optimization procedure for the optimal control.

\subsection{Stochastic gradient descent optimization}\label{SGD_SOC}
In our stochastic optimal control formulation of the SNN model, the cost function $J$ (defined in \eqref{Control-Cost}) is the expectation of the loss function $\Phi(X_T, \Gamma)$, where $\Gamma$ is the random variable corresponding to training data. 
The main theme of the stochastic gradient descent (SGD) optimization is that instead of using the average of all the data samples as a Monte Carlo approximation for the expectation, we can randomly select one data sample from the data set to approximate the expectation in each gradient descent iteration step. Following this methodology, the gradient descent iteration \eqref{Semi_Approx-GD} that we use to search for the optimal control can be achieved through the following scheme
\begin{equation}\label{SGD-Data}
u_n^{k+1} = u_n^k - \eta_k \E\big[f'_u(X_{n}, u_{n}^k)^T Y^k_n + g_u'(u_{n}^k)^T Z^k_n\big]\big|_{\gamma^k \in \{\gamma_q\}_{q=1}^Q}, \quad n = 0,1, \cdots, N-1,
\end{equation}
where $\gamma^k$ is a data randomly selected from the entire training data set $\{\gamma_q\}_{q=1}^Q$ at each iteration step $k$, and the random selection $\gamma^k$ appears in the side condition of the adjoint BSDE \eqref{Intro:BSDE}, i.e. $Y_T = \Phi'_x(X_T, \ \gamma^k)$. 
On the other hand, although we can replace the data variable $\Gamma$ by its random representation $\gamma^k$ in the SGD optimization, we still need to keep the expectation in \eqref{SGD-Data} since the system also contains randomness caused by the Brownian motion $W$ in the state process $X$, which is used to model the artificial noises in the SNN model.
In this way, the fully calculated gradient descent iteration \eqref{Full_Approx-GD} can be simplified as
\begin{equation}\label{Semi_Approx-SGD}
u_n^{k+1} = u_n^k - \eta_k \sum_{m=1}^M \frac{f'_u(X_{n}^{m, k}, u_{n}^k)^T Y^{k}_n|_{\gamma^k}(X_{n}^{m, k}) + g_u'(u_{n}^k)^T Z^k_n|_{\gamma^k}(X_{n}^{m, k})}{M }.
\end{equation}

We can see from the above SGD iteration scheme that we only need to solve the BSDE \eqref{Intro:BSDE} once to get an update for the estimated optimal control --- instead of solving the entire BSDE system $Q$-times (corresponding to different selection of data $\gamma_q$) in each iteration step to get the complete Monte Carlo approximation for the expectation. However, the major computational challenges of large-number Monte Carlo simulations and high dimensional function approximation for solutions $Y$ and $Z$ still remain.
A novel concept that we want to introduce in this work is that the noises in the SNN model can also be considered as a source of data. In other words, we treat the simulated sample paths for the controlled state process $X$ as ``pseudo-data'' that we use to calculate expectations in the Monte Carlo approximation.
Therefore, the methodology of single-sample representation for random variables in SGD optimization can be extended and applied to our computational framework to describe the gradient process, which contains both the data variable $\Gamma$ and the state process $X$.

Following this methodology, we generate a simulated sample-path for the state process at each iteration step $k$ as
\begin{equation}\label{sample-X}
X_{n+1}^k = X_{n}^k + f(X_n^k, u_{n}^k) h + g(u_{n}^k) \sqrt{h} \omega_n^k, \qquad 0 \leq n \leq N-1.
\end{equation}
Corresponding to the sample-path $\{X_n^k\}_n$ and the randomly selected data $\gamma^k$ from the training data set, we also introduce the following \textit{path-wise} numerical schemes to implement the schemes \eqref{Semi-BSDE} for the BSDE 
\vspace{-0.3em}
\begin{equation}\label{sample-BSDE}
\hat{Y}_n^k = \hat{Y}_{n+1}^k + h f_x'(X_{n+1}^k, u_{n+1}^k) \hat{Y}_{n+1}^k, \qquad \hat{Z}_n^k = \frac{\hat{Y}_{n+1}^k \sqrt{h} \omega_n^k}{h},
\vspace{-0.3em}
\end{equation}
where expectations in the schemes \eqref{Semi-BSDE} are represented by single-realization of samples indexed by the iteration step $k$, and $\hat{Y}_n^k$ and $\hat{Z}_n^k$ are approximated solutions for $Y$ and $Z$ corresponding to the state sample $X_{n}^k$, i.e. $\hat{Y}_n^k \approx Y_{t_n}(X_n^k)$ and $\hat{Z}_n^k \approx Z_{t_n}(X_n^k)$. In this way, at each gradient descent iteration step, we generate a sample-path $\{X_n^k\}_n$  for the state process and then solve the BSDE along this sample path with the schemes \eqref{sample-BSDE} to get a pair of simulated paths $\{( \hat{Y}_n^k, \ \hat{Z}_n^k)\}_n$.  
Then, with simulated sample paths obtained by \eqref{sample-X}-\eqref{sample-BSDE}, we introduce the following SGD optimization scheme
\begin{equation}\label{scheme:SGD}
u_n^{k+1} = u_n^{k} - \eta_k \Big( f'_u(X_{n}^k, u_{n}^k)^T \hat{Y}^{k}_n+ g_u'(u_{n}^k)^T \hat{Z}^k_n \Big), 
\end{equation}
where $f'_u(X_{n}^k, u_{n}^k)^T \hat{Y}^{k}_n+ g_u'(u_{n}^k)^T \hat{Z}^k_n$ is the single-sample representation for the gradient, which is under expectation.
As a result, we obtain our estimated optimal control $\{ \hat{u}_n\}_n :=  \{ u_{n}^{K} \}_n$.

\vspace{0.5em}

We notice that in the schemes \eqref{sample-BSDE}, we only use one sample of each random variable to approximate conditional expectations instead of using the Monte Carlo average of a set of samples, and the simulated paths $\{\hat{Y}_n^k\}_n$ and $\{\hat{Z}_n^k\}_n$ only describe solutions of the adjoint BSDE \eqref{Intro:BSDE} corresponding to a given simulated state sample path $\{X_n^k\}_n$, which are not complete numerical approximations for $Y$ and $Z$ in the state space. However, we need to point out that the solutions $Y$ and $Z$ of the adjoint BSDE are used to formulate the gradient process, and hence the conditional expectations in the numerical schemes \eqref{Semi-BSDE} are also used to compose the approximated gradient process. In this connection, the justification for using one sample to represent a random variable in the SGD optimization can also be applied to explain the single-sample representation for the gradient process in \eqref{scheme:SGD}.
Although we are aware that a complete description for the state variable $X$ is necessary to accurately characterize the solutions $Y$ and $Z$ since they are both functions of $X$, it is important to emphasize that the purpose of solving the optimal control problem \eqref{Controlled-State} - \eqref{Optimal-Control} is to find the optimal control $\bar{u}$, which is equivalent to the optimal parameters in the SNN model \eqref{Intro:SNN}, and it's not necessary to obtain accurate numerical solutions for $Y$ and $Z$.

\subsection{Summary of the algorithm}\label{Summary_Algorithm}

We summarize the numerical algorithm of our SMP approach for SNNs  in \textit{Algorithm \ref{Algorithm 1}.} 
\begin{table*}\caption{Numerical implementation of the SMP approach for SNNs}\label{Algorithm 1}
\vspace{0.5em}
\centering
\begin{tabular} {p{0.9\textwidth}} 
\hline\noalign{\smallskip}
{\bf Algorithm 1.}  \\
\noalign
{\smallskip}\hline
\noalign{\smallskip}
\vspace{-0.2cm}
\begin{spacing}{1.1}
\begin{algorithmic}\label{algorithm}
\item[Formulate] the SNN model \eqref{Intro:SNN} as the stochastic optimal control problem \eqref{Controlled-State} - \eqref{Optimal-Control} and give a partition $\Pi^N$ to the control problem as the depth of SNN.
\item[Choose] the number of SGD iteration steps $K \in \mathbb{N}$,  the learning rate $\{\eta_k\}_k$ and the initial guess for the optimal control $\{u_n^0\}_n$
\item[\textbf{for}] SGD iteration steps $k =0, 1, 2, \cdots, K-1$,\\
\begin{description}
\item  Simulate one realization of the state process $\{X^{k}_{n}\}_{n}$ through the scheme \eqref{sample-X}.
\item  Simulate one pair of solution paths $\{\big( \hat{Y}^{k}_{n}, \ \hat{Z}^k_n \big)\}_{n}$ of the adjoint BSDEs system \eqref{Intro:BSDE} corresponding to $\{X^{k}_{n}\}_{n}$ through the schemes \eqref{sample-BSDE};
\item  Calculate the gradient process and update the estimated optimal control $\{u^{k+1}_n\}_n$ through the SGD iteration scheme \eqref{scheme:SGD};
\end{description}
\item[\textbf{end for}]
\item[-] The estimated optimal control is given by $\{ \hat{u} _n\}_n=  \{ u_{n}^{K} \}_n$;
\end{algorithmic} 
\vspace{-2em}
\end{spacing}\\
\hline
\end{tabular}
\end{table*}
In the computational framework of our SMP approach, the SGD optimization schemes \eqref{sample-X}-\eqref{scheme:SGD} provide a feasible algorithm to search for the optimal control in the stochastic optimal control problem, therefore provide an optimization algorithm to determine the optimal parameters in the SNN model \eqref{Intro:SNN}. Since we use only one realization of simulated process $\{X^{k}_{n}\}_{n}$ to represent the state $X$ and calculate the simulated solution paths $\{\hat{Y}^{k}_{n}\}_{n}$ and $\{\hat{Z}^{k}_{n}\}_{n}$ of the adjoint BSDE corresponding to the state path $\{X^{k}_{n}\}_{n}$ at each SGD iteration step, we could avoid simulating large number of Monte Carlo samples in approximating expectations. At the same time, by using solution paths $\{\hat{Y}^{k}_{n}\}_{n}$ and $\{\hat{Z}^{k}_{n}\}_{n}$ to represent $Y$ and $Z$ in the gradient process, we don't need to obtain numerical solutions for the adjoint BSDE in the entire state space. Therefore, our SGD optimization is an efficient alternative to the fully-calculated gradient descent approach.

\section{Numerical experiments}\label{Numerics}

In this Section, we present three examples to demonstrate the performance of our SMP approach for SNNs. For all three examples, SNNs are used to model uncertainties in probabilistic machine learning. The SMP method that we developed in this work aims to solve the stochastic optimal control problem corresponding to the SNN model \eqref{Intro:SNN}, which is equivalent to the training for SNNs. The trained SNNs will then quantify the uncertainties of probabilistic models that we try to learn through neural networks. In the first example, we solve a stochastic version of classification problem, in which the classification function that separates different types of points is a noise perturbed curve.  In the second example, we let SNNs learn a function curve by using scattered function values as training data, which can be considered as a machine learning approach for function approximation. Instead of approximating a deterministic function, which can be done by using standard deterministic DNNs, we approximate stochastic functions. In both the classification problem and the function approximation problem, we compare our SMP method with the Bayesian Neural Network (BNN), which is the state-of-the-art  probabilistic learning method, to demonstrate the effectiveness and efficiency of our algorithm, and the optimization procedure in the BNNs is implemented by Markov Chain Monte Carlo (MCMC) sampling. In the third example, we use SNNs to solve a parameter estimation problem, in which both the perturbation noises in the parameterized model and the uncertainties in data collection are considered. In all the numerical examples that we carry out in this section, we use the sigmoid function as our activation function with weights and biases as parameters, and we choose the temporal discretization step-size $h=1$. All the parameters are initialized by standard multi-variate Gaussian variables for the BNN, and we choose the mean of the initial guess for weights as $0$, and we choose the mean of the initial guess for biases as $0.05$. For the SMP approach, we let the weights be initialized by random numbers drew from a standard multi-variate Gaussian distribution. The initial guess for biases is chosen as $0.05$, and the initial guess for the diffusion coefficients is chosen as $0.01$. The learning rate is chosen as $\eta_k=\frac{1}{\sqrt{k}}$, where $k$ is the number of iteration step.


\subsection{Classification}
In this example, we use SNNs to solve a stochastic version of classification problem, which is a benchmark application of DNN. The goal of classification problems is to assign input points different labels from a fixed set of categories. When solving a classification problem by machine learning, we use scattered classified points as training data to train a DNN that will assign test points appropriate classification labels \cite{Learning-Classification}. This can also be considered as using training data to learn a classification function, which separates a domain into several regions. The classification function $f$ that we choose in this example is a circle in the $xy$-plane, i.e. 
$$f(r, \ \theta) = r (\sin \theta, \ \cos \theta), \quad \text{with} \quad r >0, \ \theta \in [0, 2 \pi].$$ 
We let the points inside of the circle be classified as one type of points (marked by red) and assign them the classification weight $0$; we let the points outside of the circle be classified as the other type of points (marked by blue) and assign them the classification weight $1$. In order to  validate the necessity of applying SNNs to stochastic classification and examine the performance of our SMP method, we allow the classification function to be perturbed by a random noise $\xi$, i.e. 
$$f(r, \ \theta, \ \xi) = (r + \xi) (\sin \theta, \ \cos \theta),$$
where $\xi$ is as a Gaussian random variable with standard deviation $0.1 r$. In this way, the classification function that we use to define the boundary of two regions becomes a random variable whose mean curve coincides the deterministic classification function $f(r, \ \theta)$.
\begin{figure}[h!]
\begin{center}
\subfloat[Samples classified by deterministic classification function]{\includegraphics[scale = 0.5]{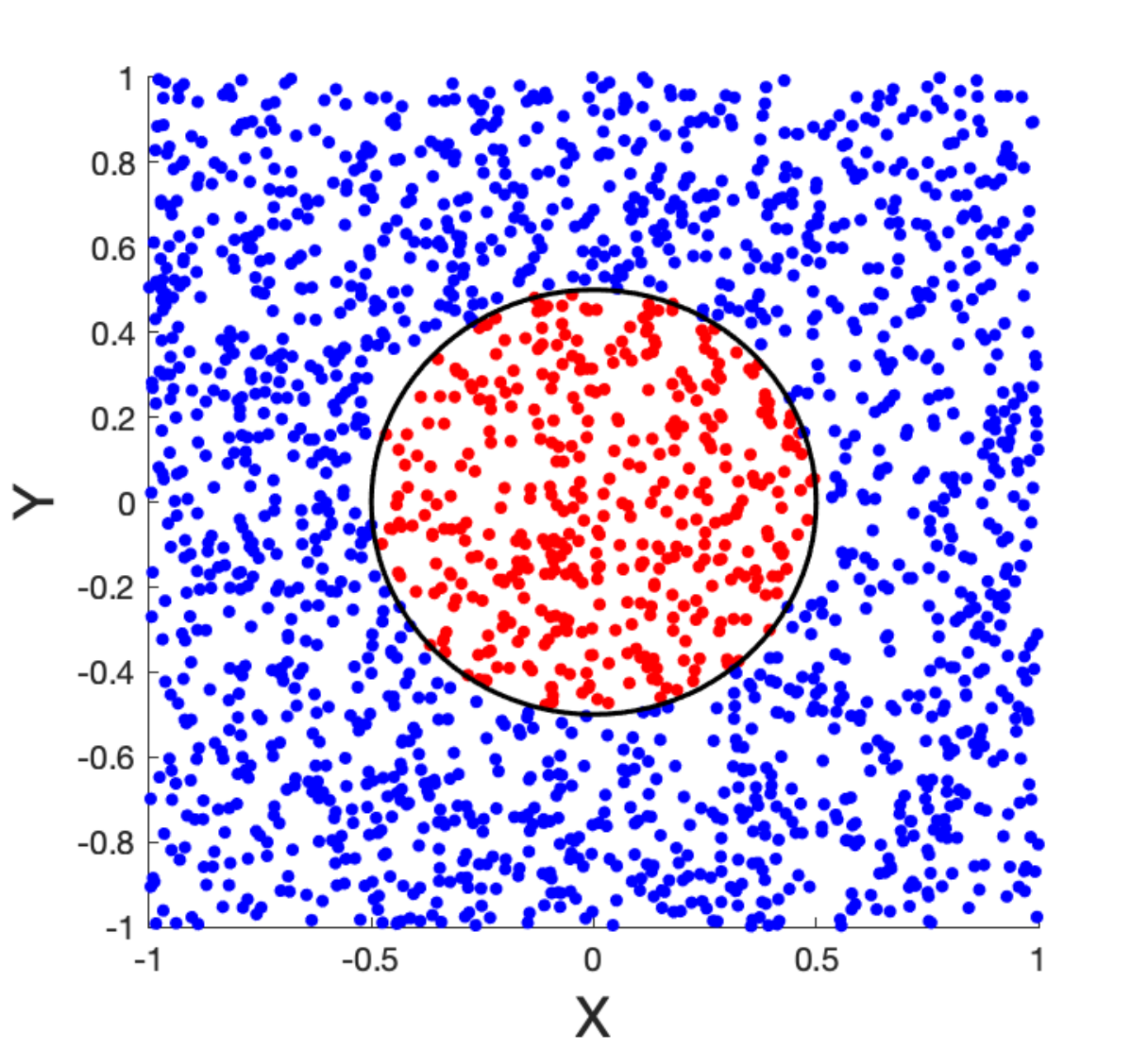} } \qquad
\subfloat[Samples classified by random classification function]{\includegraphics[scale = 0.5]{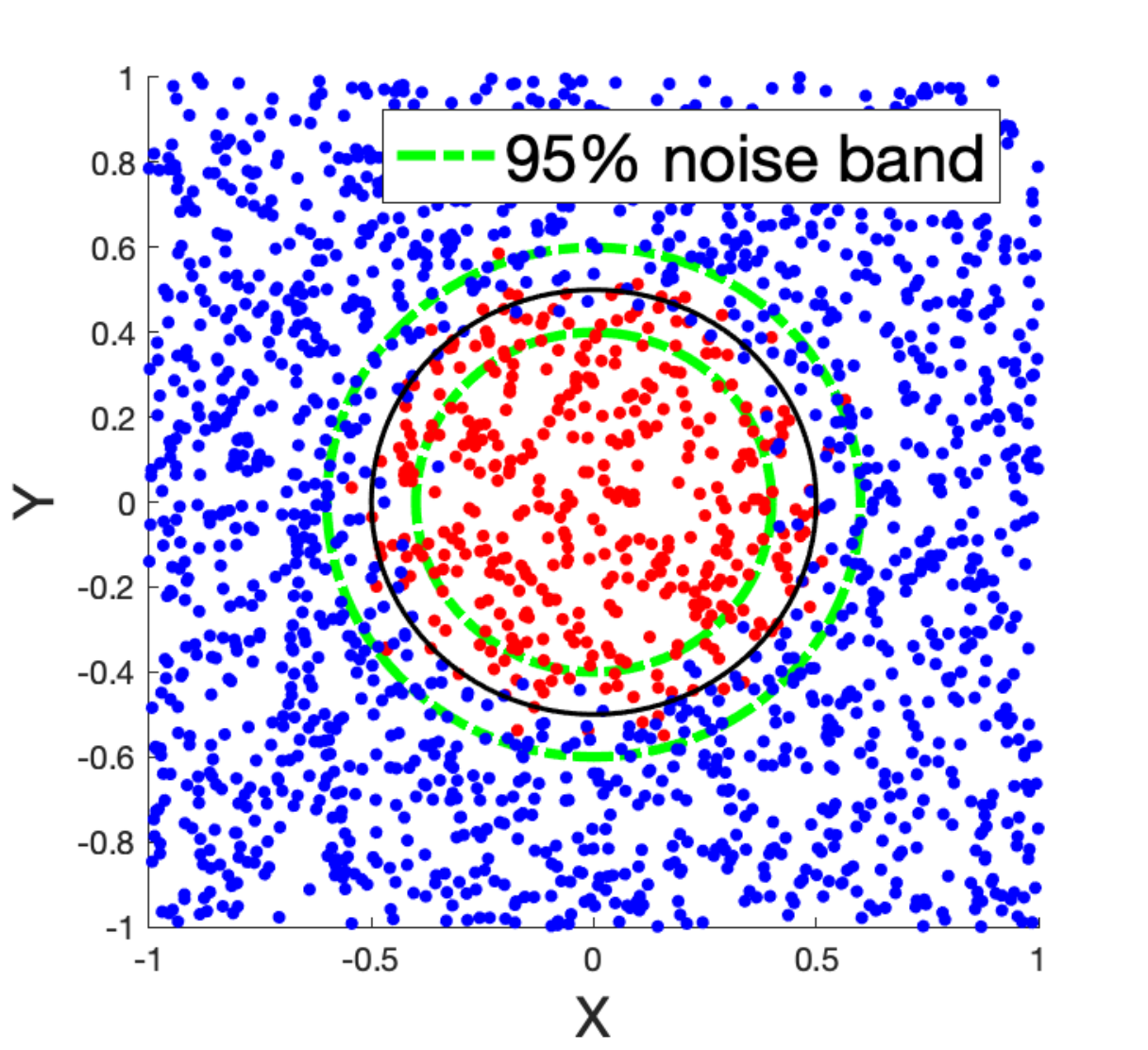} }\end{center}
\caption{Classified samples of the deterministic and random classification functions. }\label{Ex1_Data} 
\end{figure}
In Figure \ref{Ex1_Data} (a), we let $r = 0.5$ and plot examples of classified samples with respect to the \textit{deterministic} classification function $f(r, \ \theta) $;  in Figure \ref{Ex1_Data} (b), we present examples of classified samples with respect to the \textit{random} classification function $f(r, \ \theta, \ \xi)$. The classification domain in this example is chosen to be a square $[-1, 1] \times [-1, 1]$. We can see from Figure \ref{Ex1_Data} (a) that all the samples inside of the circle are classified as red and samples outside of the circle are classified as blue. On the other hand, we can see from Figure \ref{Ex1_Data} (b) that when using the random classification circle $f(r, \ \theta, \ \xi)$, samples outside of the circle are occasionally classified as red and a few samples inside of the circle are classified as blue. In Figure \ref{Ex1_Data} (b), 
we also plot green dashed circles to describe the $95\%$ noise band (the $2\sigma$ region) caused by the random variable $\xi$, and we can see from this subplot that almost all the mis-classified samples are within the $95\%$ noise band.

\begin{figure}[h!]
\begin{center}
\subfloat[BNN classification results]{\includegraphics[scale = 0.5]{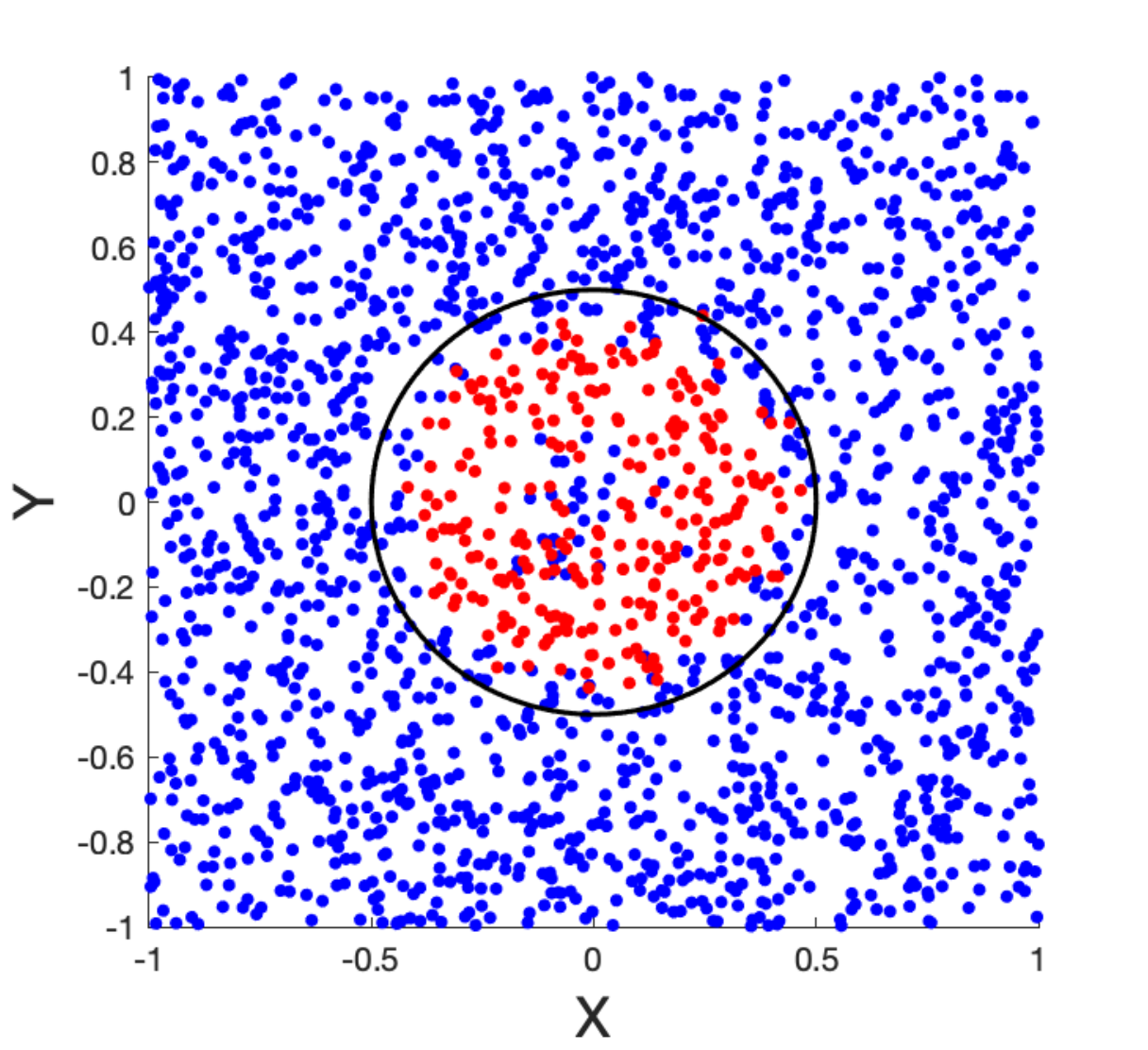} }  \qquad
\subfloat[SMP classification results]{\includegraphics[scale = 0.5]{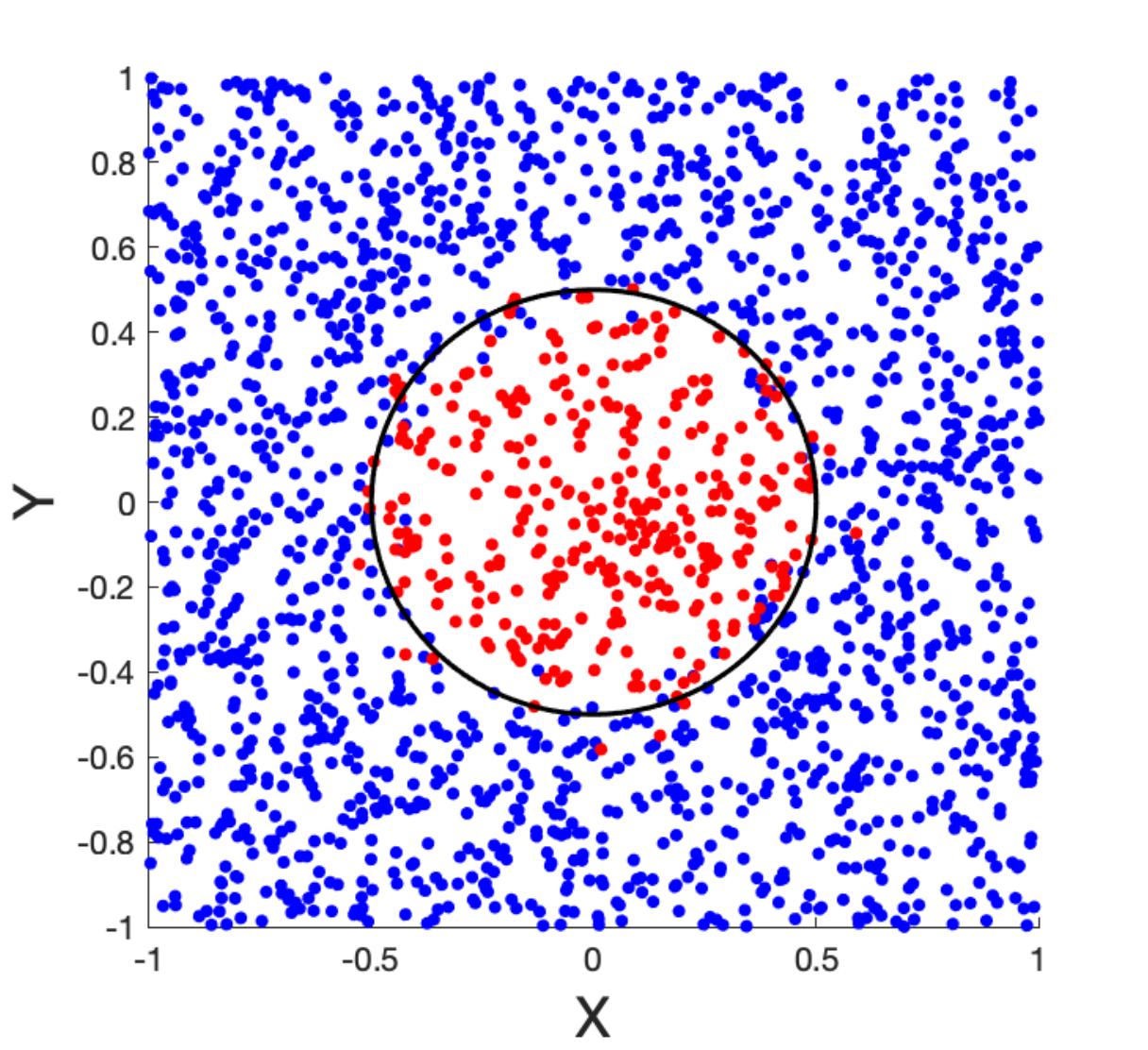} }
\end{center}
\caption{Comparison between BNN output and SMP output. }\label{Ex1_Comparison} 
\end{figure}

Since the classification function that we use to classify samples is a random variable, standard deterministic DNNs are not suitable. In this example, we use probabilistic machine learning to learn the random classification function.  In Figure \ref{Ex1_Comparison}, we compare the performance of the BNN approach with our SMP method for the SNN model. Both the BNN and the SMP method use moderate size neural networks (with $2$ neurons and $8$ layers) to model the classification action, and the classification criteria to distinguish red and blue as output is $0.5$. We run $10^{6}$ training steps in the BNN approach due to the large number of parameter variables to be sampled in Bayesian inference, and we use $10^5$ iteration steps in the SGD optimization procedure in the SMP approach. The black solid circle in each subplot represents the deterministic classification function $f(r, \ \theta)$ (without the perturbation $\xi$) as a reference boundary for classification results. 
In Figure \ref{Ex1_Comparison} (a), we plot the classification results of $2,000$ samples classified by the BNN approach and Figure \ref{Ex1_Comparison} (b) gives the classification results of $2,000$ samples classified by the SMP method. We can see from this figure that the BNN fails to provide a proper classification boundary -- even using more training steps, and there are several mis-classified samples around the center of the circle caused by the randomness of training data. On the other hand, our SMP method not only accurately captured the shape of classification circle, the classified samples around the classification boundary also have very similar scattered behavior compared to the original data presented in Figure \ref{Ex1_Data} (b).

\begin{figure}[h!]
\begin{center}
\includegraphics[scale = 0.65]{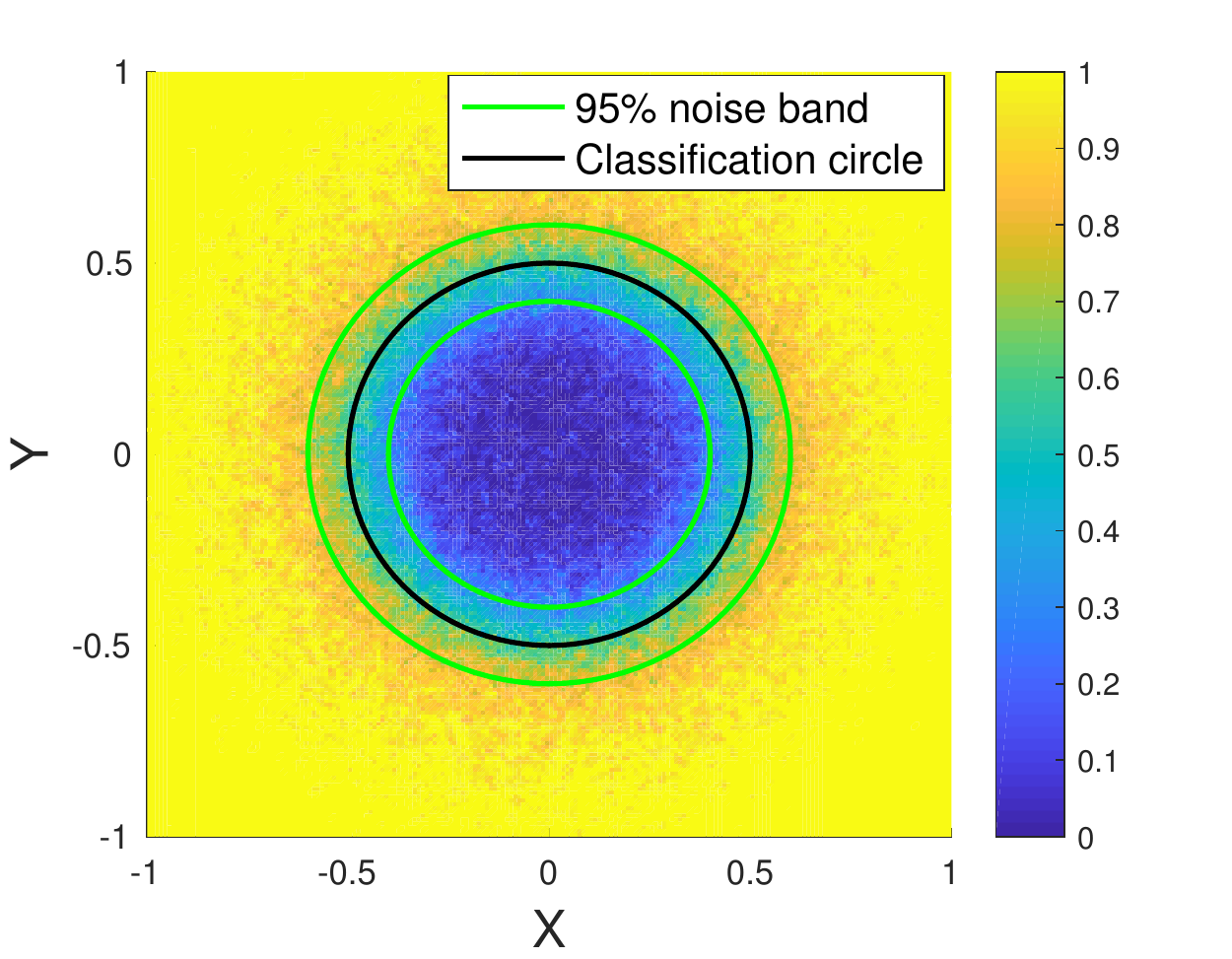}
\end{center}\vspace{-0.5em}
\caption{Weight surface for classification results obtained by SMP. }\label{Ex1_Distribution} 
\end{figure}

As a probabilistic learning model, the trained SNN always gives random output. Therefore, simply comparing $2,000$ classification results obtained by using our SMP method with the true classified data does not provide enough evidence for the uncertainty quantification performance of SNNs.
To present more detailed behavior of SNN output, we draw the weight surface approximated by $10^5$ test samples classified by our SMP trained SNN in Figure \ref{Ex1_Distribution}. The approximation method that we use to approximate the weight surface is the radial basis function approximation, which gives a smooth interpolatory approximation by using the weights on classified test samples. 
The blue part of the weight surface is corresponding to the classification weight $0$ and the yellow part of the weight surface is corresponding to the classification weight $1$. The deterministic classification circle and the $95\%$ noise band (caused by the perturbation $\xi$) are plotted by the solid black circle and solid green circles, respectively. We can see from this figure that the weight surface obtained by using our SMP method clearly separates the inside/outside of the circle. Moreover, the weight surface near the classification circle has fuzzy features, which reflect the uncertainty of the SNN output; and the fluctuation region is well-aligned with the $95\%$ noise band for the stochastic classification function, which explains the similarity of random classification results near the classification boundary between the SNN output  (in Figure \ref{Ex1_Comparison} (b)) and the original data (in Figure \ref{Ex1_Data} (b)).




\subsection{Function approximation}

In this example, we use SNNs to approximate stochastic functions. Function approximation is an important application of artificial neural networks. By using the independent variable of a function as input and its corresponding function values as data, we can train a neural network that models the mapping between the independent variable and function values. As an alternative method to classic approximation techniques, which approximate a target function by basis functions,  neural networks can be considered as meshfree universal approximators \cite{Learning-approximation},
which are potentially capable to describe complicated high dimensional functions. When approximating a stochastic function, SNNs are needed to model the uncertainty of the function. 

The first function that we are going to approximate is a cubic function on the interval $[0, 1]$, which is perturbed by a spatial-invariant noise $\xi$, i.e. 
\begin{equation*}
f(x,\ \xi) = 2 + (1+x)^3 + \xi.
\end{equation*}
The deterministic part of the function gives a deterministic curve and $\xi$ is a Gaussian random variable that brings noises to function values, and we choose the standard deviation for $\xi$ as $0.2$. In this way, each data that we collect to train the neural network is a pair $\big(x^i, f(x^i, \xi^i) \big)$, where the input $x^i$ is randomly selected from $[0, 1]$,  $\xi^i$ is a sample of $\xi$, i.e. $\xi^i \sim \xi$, and the size of the noise is independent from the selection of point $x^i$. 
\begin{figure}[h!]
\begin{center}
\includegraphics[scale = 0.9]{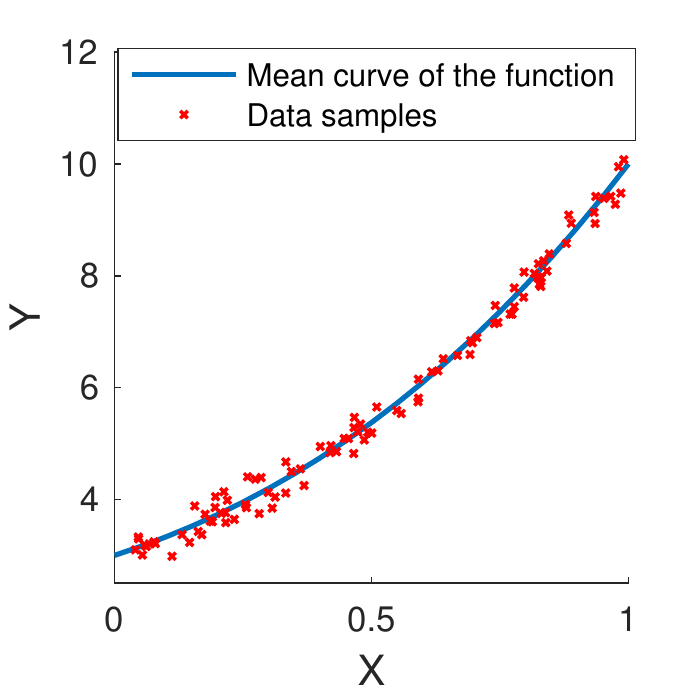}
\end{center}
\caption{Example of data samples.}\label{Ex2_data_case1} 
\end{figure}
In Figure \ref{Ex2_data_case1}, we plot $100$ data samples by using red crosses to demonstrate the randomness of training data, and the mean curve of the stochastic function $f(x,\ \xi)$ is represented by the solid blue curve. Although the function we approximate is one-dimensional, the model we try to capture gives continuous mapping from input to output, which is more complicated then simple classifications. Moreover, since the target function that we want to approximate is perturbed by a random noise $\xi$, the criteria to measure the performance of machine learning is not only the accuracy of approximating function's mean curve, but also the capability of estimating the uncertainty bound of function's stochasticity. 

\begin{figure}[h!]
\begin{center}
\subfloat[Performance of the BNN approach]{\includegraphics[scale = 0.65]{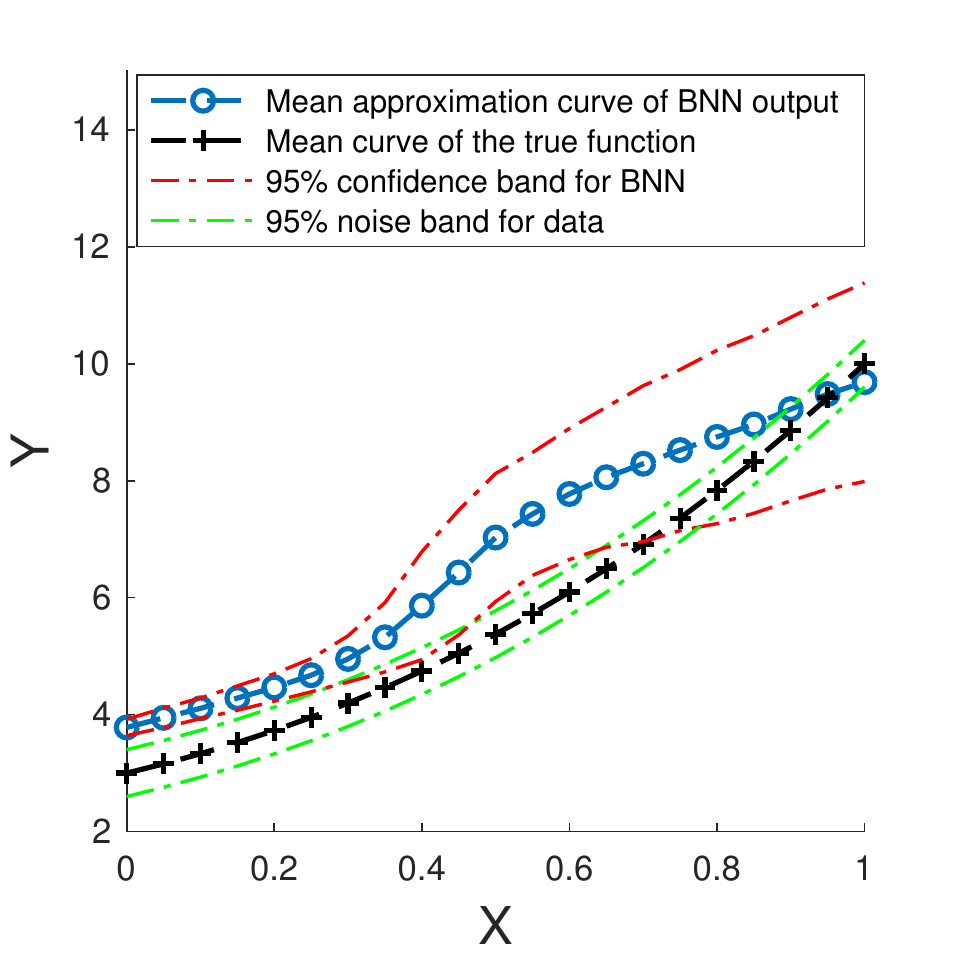} }  \qquad
\subfloat[Performance of the SMP approach ]{\includegraphics[scale = 0.65]{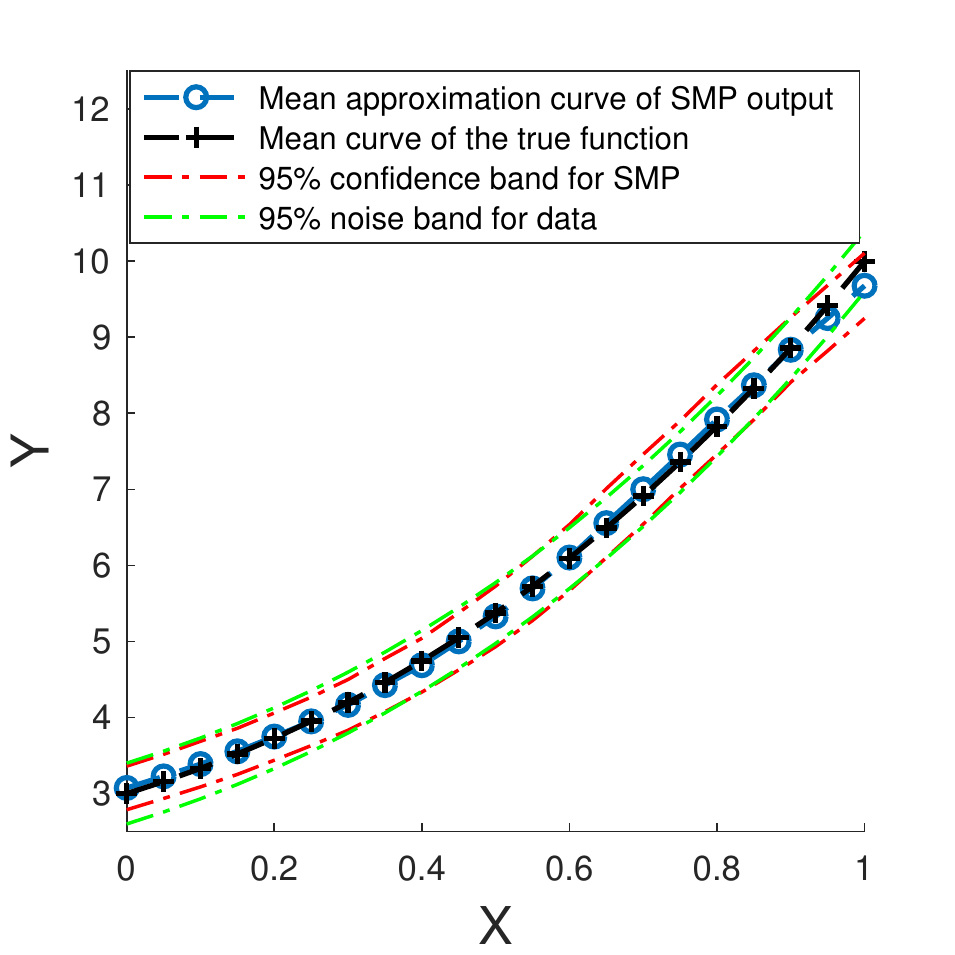} }\end{center}
\vspace{-0.5em}
\caption{Comparison of BNN approach for SNNs and SMP approach for SNNs. }\label{Ex2_Comparison_case1} 
\end{figure}

The probabilistic learning methods that we use to learn the stochastic function $f(x,\ \xi)$ in this experiment are the BNN and our SMP method. In Figure \ref{Ex2_Comparison_case1}, we compare the performance of the BNN approach with the SMP approach in training $3$-neuron, $8$-layer neural networks. 
To train the SNN with our SMP method, we use $2 \times 10^5$ SGD iteration steps to find the optimal control in the state process in our stochastic optimal control formulation of SNNs. For the BNN approach, we carry out $2 \times 10^7$ MCMC steps in the Bayesian inference (due to the high dimensionality of the parameter space to sample), which is $100$-time of the iteration number that we use in the SMP method, and we keep the last $10^6$ MCMC steps as samples to formulate the distribution for the desired BNN parameters. 
In each subplot of Figure \ref{Ex2_Comparison_case1}, we use the black curve marked by crosses to represent the mean curve of the true function and the green dashed curves to represent the $95\%$ noise band for the function due to the perturbation $\xi$.
In Figure \ref{Ex2_Comparison_case1} (a), we present the approximation performance of the BNN approach, where the blue curve marked by circles is the mean approximation curve obtained by averaging BNN output samples and the red dashed curves provide the $95\%$ confidence band of BNN output. In Figure \ref{Ex2_Comparison_case1} (b), we present the approximation performance of the SMP approach, where the blue curve marked by circles is the mean approximation curve obtained by averaging SMP trained SNN output samples and the red dashed curves provide the $95\%$ confidence band of SNN output.  From this figure, we can see that the BNN could capture the main feature of the target function by using more iteration steps. However, the confidence band of the BNN output does not accurately describe the uncertainties of the target function caused by the random perturbation $\xi$. On the other hand, the SMP method accurately captured the function's mean curve, and the confidence band provided by the SNN output is also well-aligned with the noise band of the target stochastic function, which indicates the success of SMP in training SNNs in accomplishing the uncertainty quantification task.

To further demonstrate the performance of our SMP approach for SNN, we approximate a stochastic function, which contains a space-dependent noise. Specifically, the function that we are going to approximate is defined by 
\begin{equation*}
F(x,\ \xi(x)) = \big(1 + \tan(1.3 \cdot x) \big)\big(1+ \xi(x)\big),
\end{equation*}
where $\xi(x) \sim N(0, \sigma^2)$ is a Gaussian random variable that brings proportional noises to the deterministic part of the function, and we choose the portion $\sigma$ to be $0.05$ in this experiment.
\begin{figure}[h!]
\begin{center}
\includegraphics[scale = 0.9]{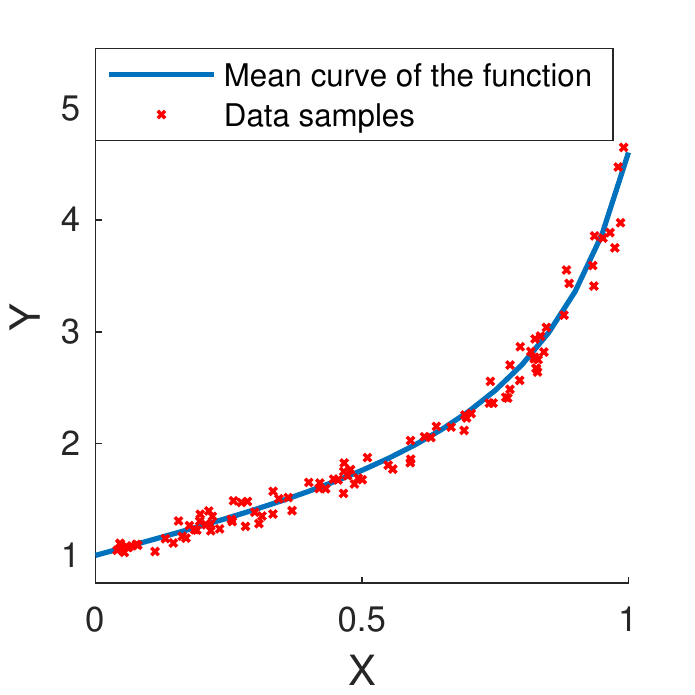}
\end{center}
\caption{Example of data samples.}\label{Ex2_data_case2} 
\end{figure}
In Figure \ref{Ex2_data_case2}, we plot $100$ data samples to demonstrate the feature of training data. We can see from this figure that when the $X$-values are getting larger, data samples generally have larger deviations from the mean curve of the function due to larger noise levels of $\xi(x)$.

\begin{figure}[h!]
\begin{center}
\subfloat[Performance of the BNN approach]{\includegraphics[scale = 0.65]{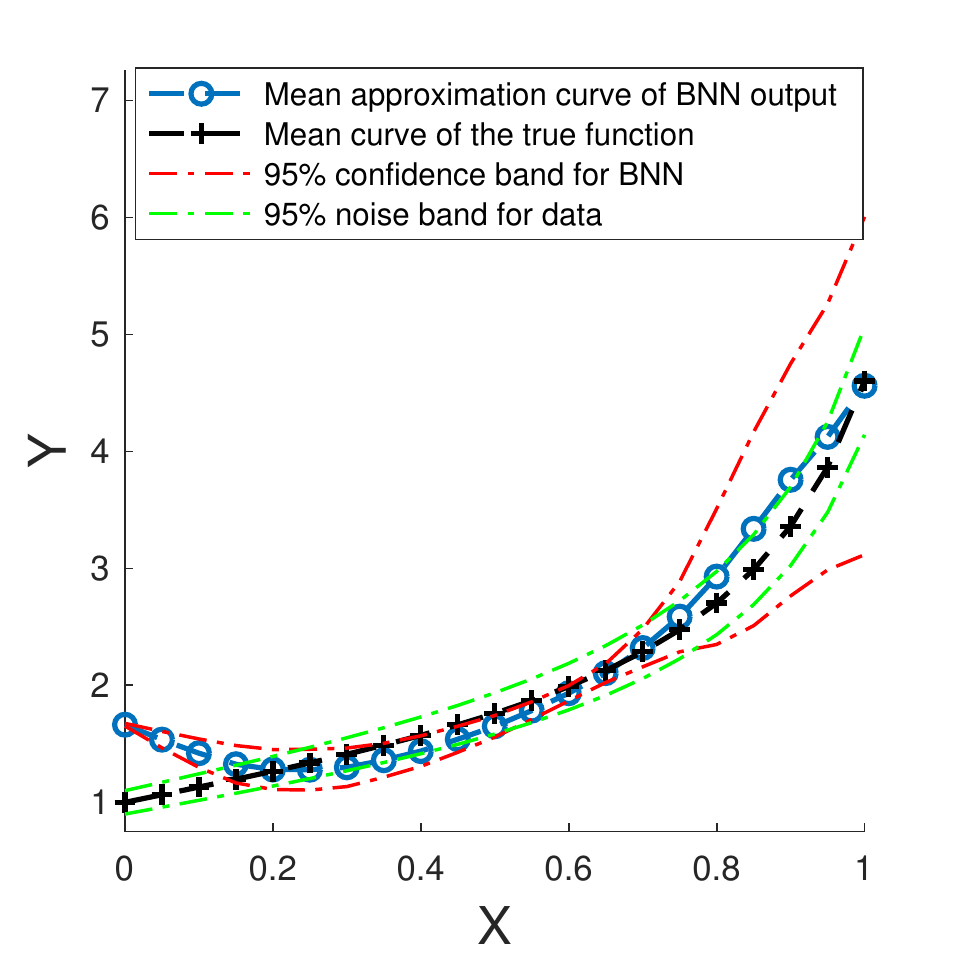} }   \qquad
\subfloat[Performance of the SMP approach]{\includegraphics[scale = 0.65]{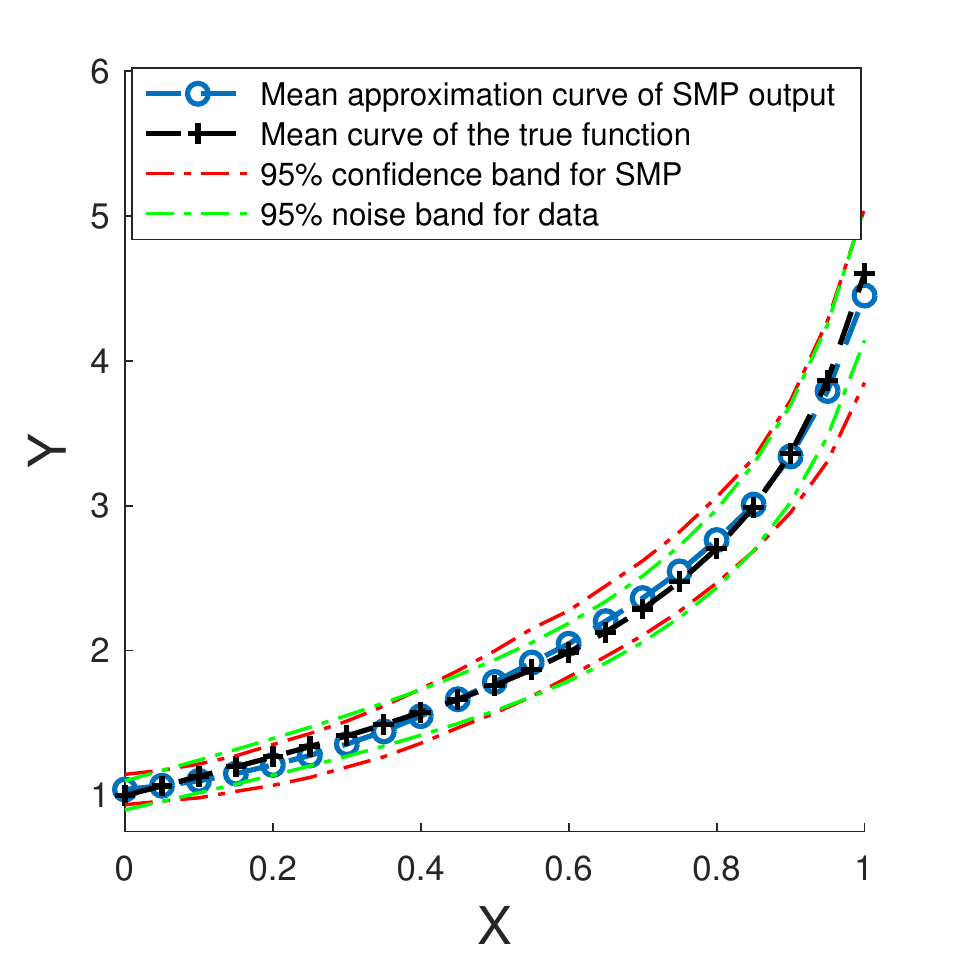} }\end{center}
\caption{Comparison of BNN approach for SNNs and SMP approach for SNNs.  }\label{Ex2_Comparison_case2} 
\end{figure}

In this experiment, we compare the performance of BNN with SMP in training $3$-neuron $12$-layer neural networks in Figure \ref{Ex2_Comparison_case2}; and we use $2\times 10^7$ MCMC steps for the BNN approach and $2 \times 10^5$ iteration steps for the SMP approach. 
The performance of the BNN approach is presented in Figure \ref{Ex2_Comparison_case2} (a) and the performance of the SMP method is presented in Figure \ref{Ex2_Comparison_case2} (b). In each subplot, the black curve marked by crosses is the mean curve of the true function and the green dashed curves give the $95\%$ noise band for the function; the blue curve marked by circles is the mean approximation curve obtained by its corresponding method (BNN for the subplot (a) and SMP for the subplot (b)) and the red dashed curves give the $95\%$ confidence band of its corresponding neural network output. We can see from this figure that although BNN provided accurate estimates for the function curve -- expect at the left end-point, the confidence band calculated from the BNN output samples can not capture the real function uncertainties.  At the same time, the SMP method captured both the function curve and the $95\%$ noise band with only $2\times 10^5$ iteration steps. In addition, Figure \ref{Ex2_Comparison_case2} (b) clearly shows that the SMP trained SNN accurately described the increasing trend of the noise band caused by the proportional noise $\xi(x)$, which indicates the effectiveness of our SMP method in quantifying the uncertainty of the model.

We want to emphasize in this function approximation example that the advantageous performance of our SMP method is based on the fact that we avoid the challenge of high dimensional Bayesian inference in the BNN approach, and we only need to solve a high dimensional point estimation problem, which is much easier than high dimensional sampling in Bayesian inference. Especially in this example, the data that we use to train the BNN are one dimensional function values, and it's typically difficult to use low dimensional data to tune MCMC samples in a very high dimensional parameter space.

\subsection{Parameter estimation}
In the third example, we examine the performance of our SMP method in solving a parameter estimation problem.  Different from traditional parameter estimation methods, which treat a parameterized function as the forward model and estimate the parameter by solving an inverse problem \cite{Morzfeld_Parameter}, the explicit expression for the parameterized model is not directly involved in the machine learning approach for parameter estimation, and people let neural networks learn the relation between the model status and its corresponding parameter from training data. Once a neural network is trained, it will produce the desired estimated parameter by using the input data of parameterized function values, which are generated by the ``hidden'' true parameter.
In this example, we use the following function 
$$
f_{\alpha}(x) = \exp(\alpha \cdot x^2/2)
$$ 
as our parameterized model, where $\alpha$ is the target parameter. 
\begin{figure}[h!]
\begin{center}
\includegraphics[scale = 0.8]{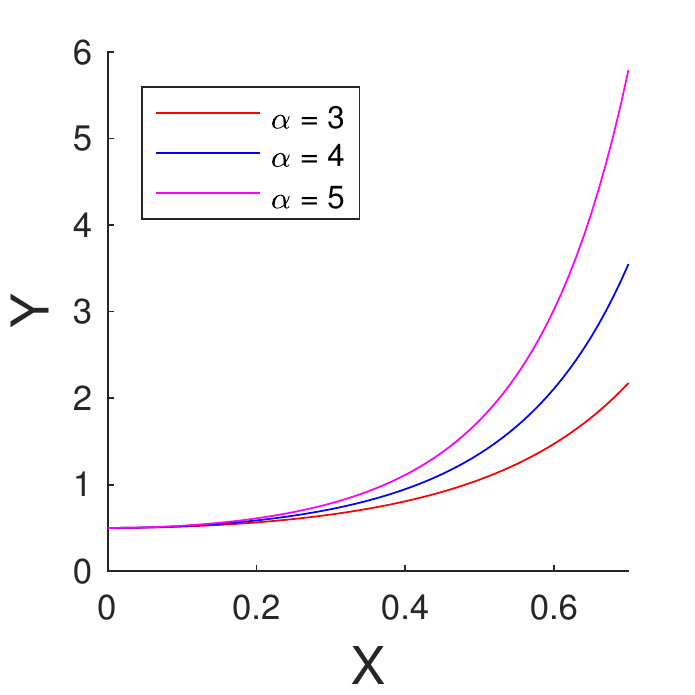}
\end{center}
\caption{Behavior of model curve with different parameters }\label{Ex3_Model}
\end{figure}
In Figure \ref{Ex3_Model}, we plot curves of  $f_{\alpha}$ with different choices of $\alpha$, i.e. $\alpha = 3, 4, 5$,  to demonstrate the influence of parameters to the parameterized model.

The training set that we use in this example are randomly selected pre-chosen parameters (from the parameter interval $\alpha \in [3, 5]$) as output, and their corresponding function values are taken as input. In practice, the data are usually perturbed by noises and there are certain facility restrictions that limit our capability to collect data. To mimic those practical scenarios, we assume that the parameterized model $f_{\alpha}$ is perturbed by a Gaussian noise with standard deviation $0.05$, and the data of parameterized function values can only be collected on spatial points $x$ that are generated by the random variable $0.5 + \gamma$, where $\gamma$ is another Gaussian variable with standard deviation $0.05$. 
Therefore, we have considered the perturbation noises in the parameterized model as well as uncertainties/limitations in data collection. In our probabilistic machine learning, we use a $3$-neuron $16$-layer SNN model to solve this parameter estimation problem, and we run $10^5$ SGD iteration steps in the SMP approach to train the SNN. In what follows, we present numerical experiments to examine the performance of our SMP trained SNN in various aspects. 

In the first experiment, we choose the true parameter to be $\alpha = 4$, and use this true parameter to generate $100$ noise perturbed function values (on randomly selected spatial points that follow the random variable $0.5 + \gamma$) as testing input data. By using only $100$ data samples as input evidence of parameter, our SMP trained SNN gives an estimated parameter $\tilde{\alpha} = 3.9453$, which is very close to the true parameter.
\begin{figure}[h!]
\begin{center}
\subfloat[Estimated function and data samples. ]{\includegraphics[scale = 0.65]{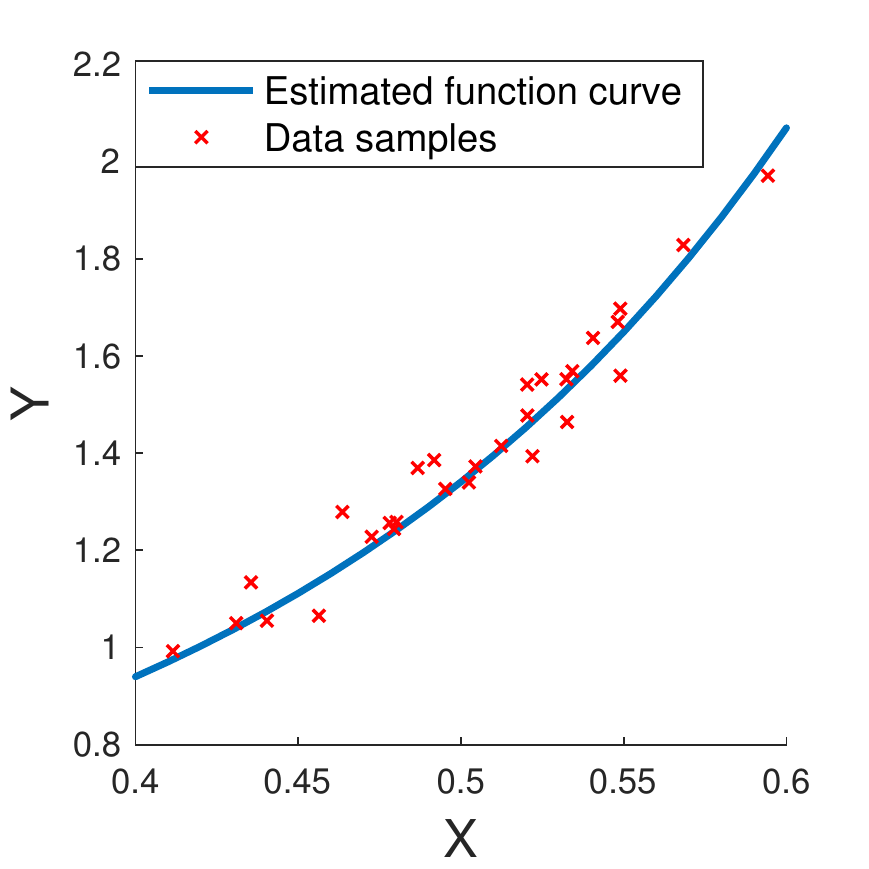} }  \qquad
\subfloat[SNN prediction for the function curve.]{\includegraphics[scale = 0.65]{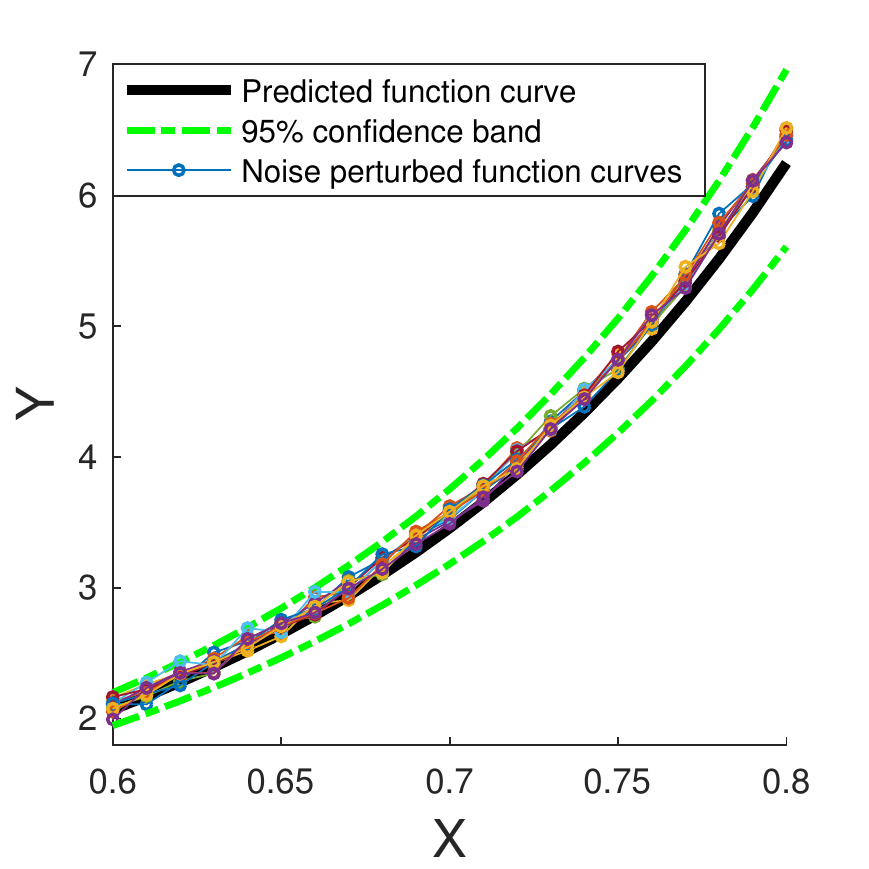} }
\end{center}
\caption{Prediction performance of SNNs for parameter estimation }\label{Ex3_Prediction} 
\end{figure}
In Figure \ref{Ex3_Prediction} (a), we plot $30$ out of $100$ samples (red crosses) to demonstrate the scattering feature of input data that we use to estimate the parameter, and we also present the estimated function curve obtained by using the estimated parameter $\tilde{\alpha}$ in $f_{\alpha}$ (the solid blue curve). From this subplot, we can see that the data samples are concentrated on the interval $[0.4, 0.6]$ with more samples near $x = 0.5$, and the estimated function curve fits the testing data well. An important purpose of parameter estimation is to predict the model behavior when direct measurements of model are not available. In Figure \ref{Ex3_Prediction} (b), we use the estimated parameter $\tilde{\alpha}$, obtained by processing $100$ testing data near $x=0.5$, to predict the function curve on the interval $[0.6, 0.8]$, where almost no data are collected. The black solid line in this subplot is the predicted function curve calculated by the estimated parameter; the green dashed curves give the $95\%$ confidence band of SNN output; and the colored curves marked by circles give $10$ realizations of noise perturbed real function curves obtained by adding perturbation noises to $f_{\alpha}$ with the true parameter. We can see that our predicted function curve fits the validation function curves well, and all the noise perturbed real function curves are within the $95\%$ confidence band, which is calculated by the SNN's distributed output.

\begin{figure}[h!]
\begin{center}
\subfloat[Output distribution for $\alpha= 3.75$]{\includegraphics[scale = 0.55]{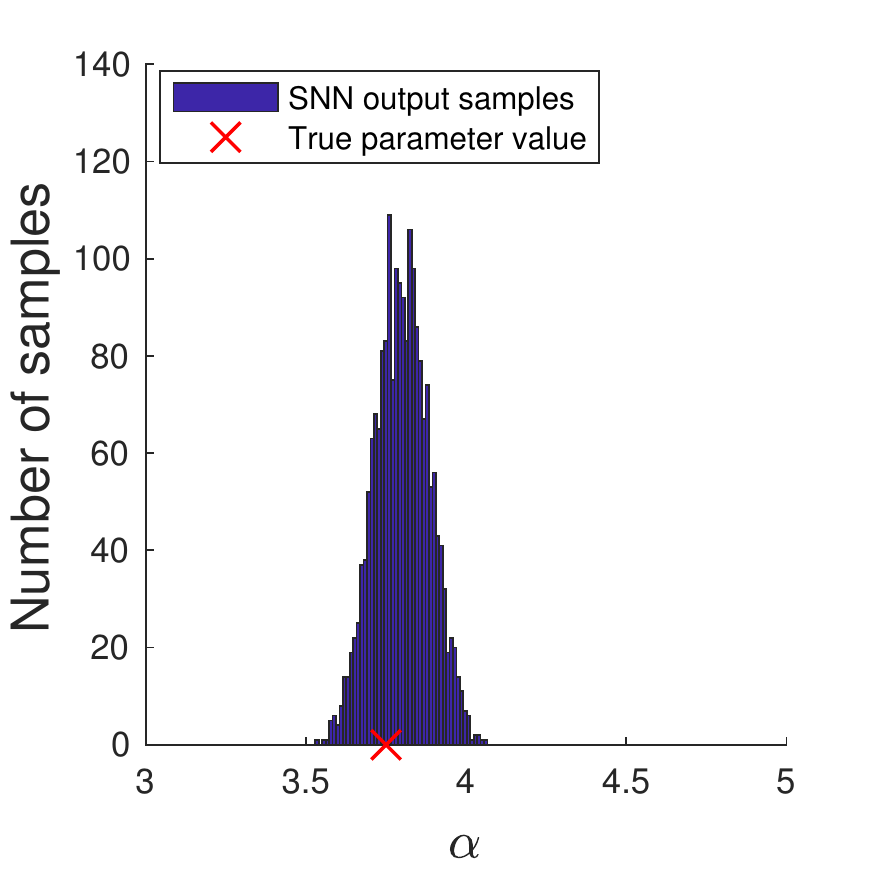} }  \
\subfloat[Output distribution for $\alpha= 4$]{\includegraphics[scale = 0.55]{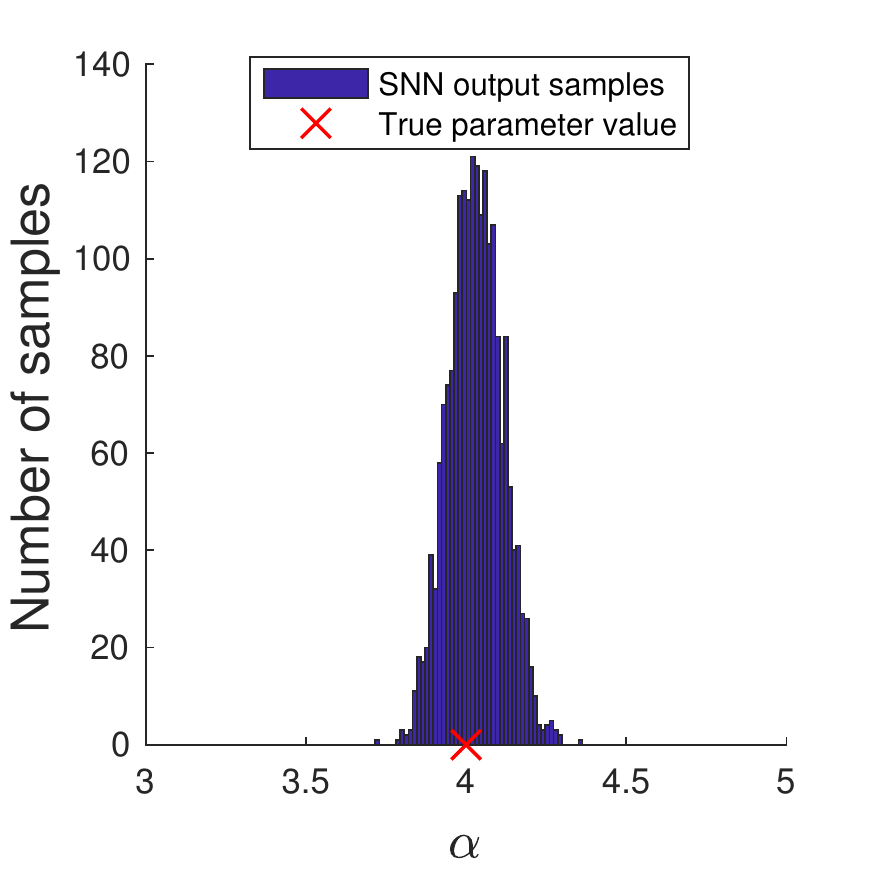} } \
\subfloat[Output distribution for $\alpha= 4.25$]{\includegraphics[scale = 0.55]{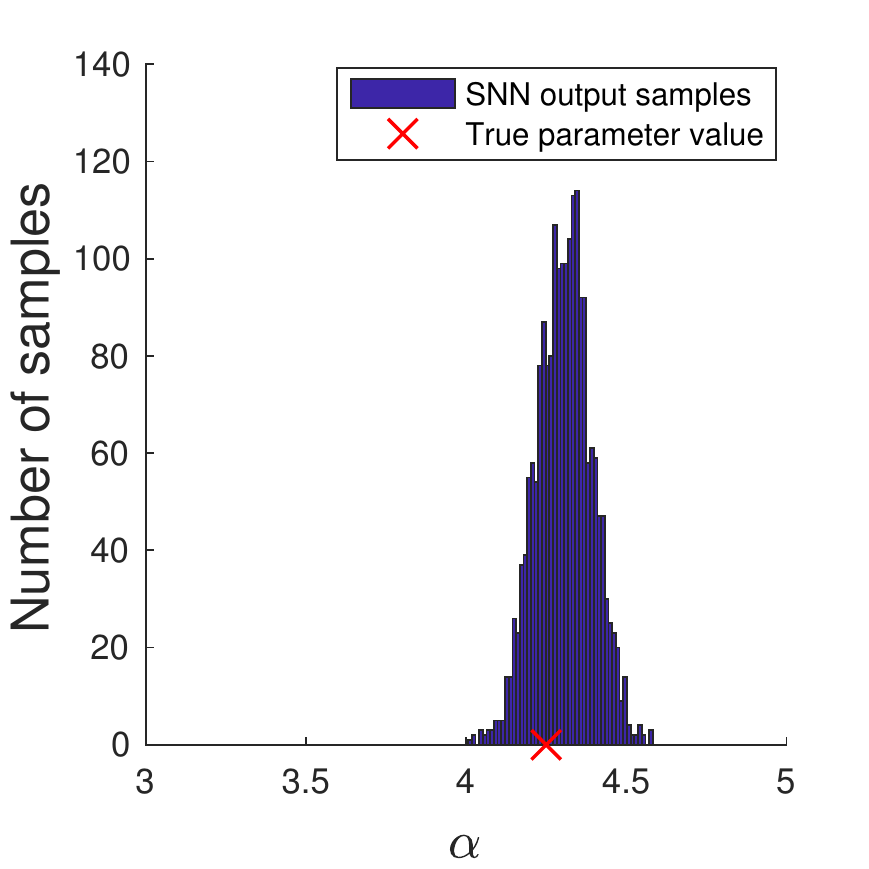} }
\end{center}
\caption{Distributions for SNN output. }\label{Ex3_Distributions} 
\end{figure}
To illustrate the probabilistic feature of SNN output, we repeat the above experiment by choosing the true parameters to be $\alpha = 3.75, 4$, and $4.25$, and plot the parameter estimation distribution of SNN output in Figure \ref{Ex3_Distributions} (a), (b), and (c), respectively, where the true parameter is given by a red cross in each subplot and each distribution is composed of $2,000$ SNN output samples. From this figure, we can see that the SNN outputs formulate effective distributions for estimated parameters and the true parameter in each case lies in the mode of SNN output distribution.

In addition to the illustration of SNN's probabilistic output in describing parameters, we shall also present an experiment to show more detailed performance of SNN in predicting model function values. To demonstrate the capability of the SMP trained SNN in recognizing data that are generated by different parameters, we screen parameters over the parameter interval $\alpha \in [3, 5]$ and use each parameter to generate $100$ noise perturbed function values on spatial points (that follow the random variable $0.5 + \gamma$) as input data.  Then, we use our SMP trained SNN to process different sets of input data (corresponding to different parameters) and generate SNN output as estimated parameter samples. 
\begin{figure}[h!]
\begin{center}
\subfloat[Predicted function values corresponding to different parameters at $x = 0.4$. ]{\includegraphics[scale = 0.6]{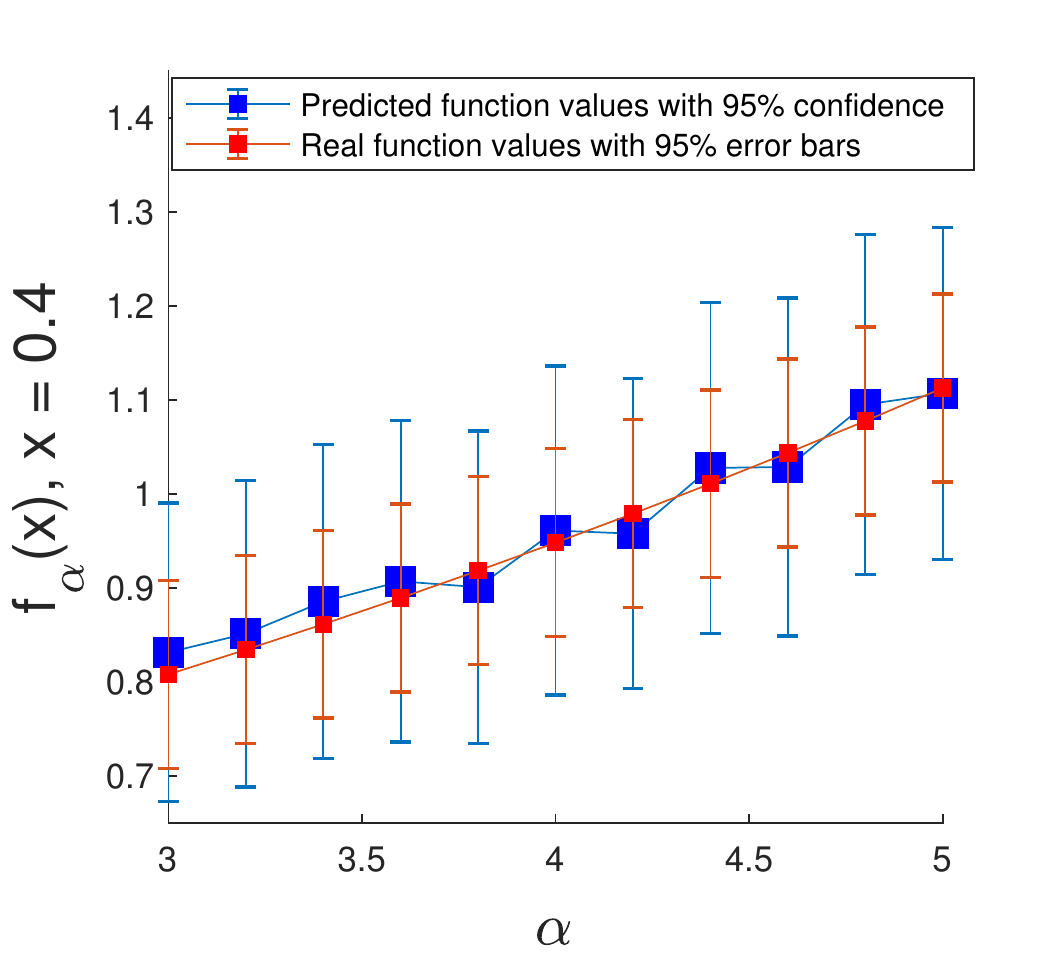} } \qquad
\subfloat[Predicted function values corresponding to different parameters at $x = 0.6$.]{\includegraphics[scale = 0.6]{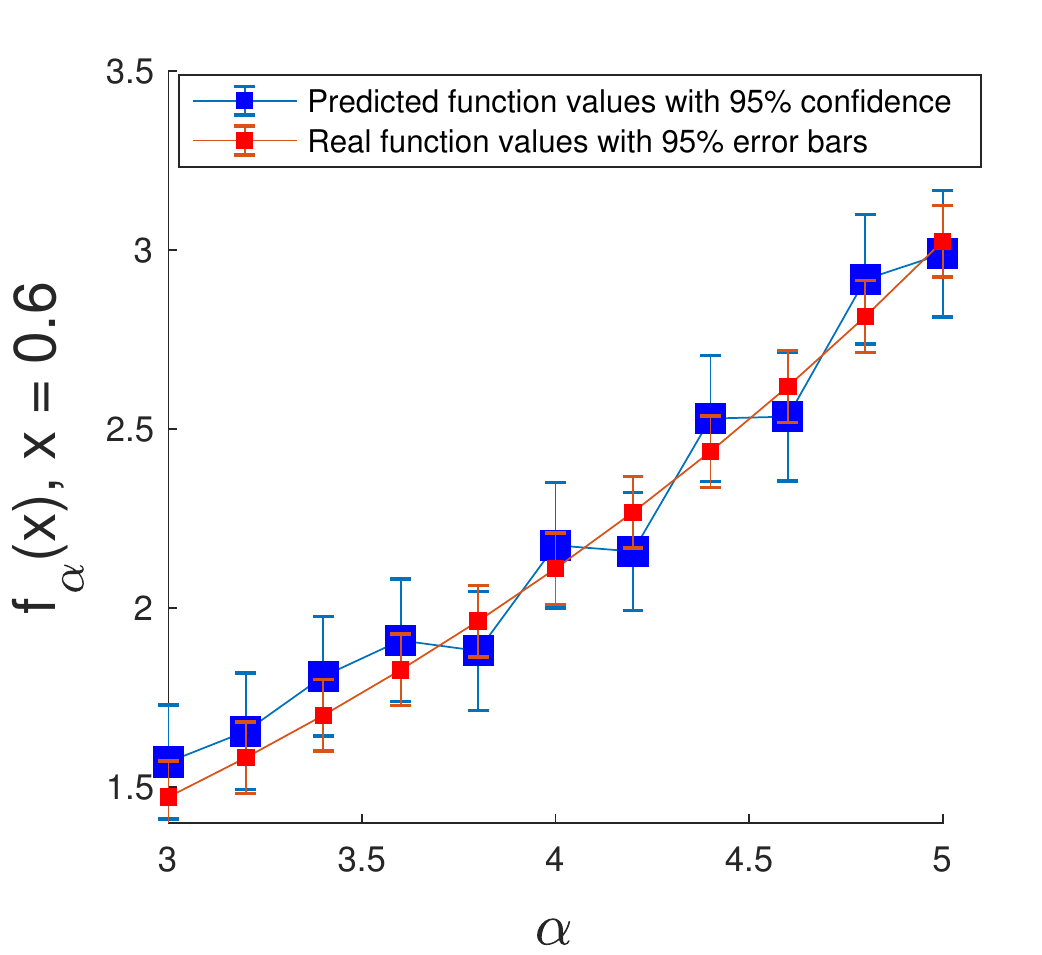} }\end{center}
\caption{Predicted function values corresponding to different parameters. }\label{Ex3_PointPrediction} 
\end{figure}
To verify the accuracy of parameter estimation by SNN, we use the estimated parameters to predict function values at spatial points $x = 0.4$ and $x = 0.6$, where function values are barely observed, and we compare our predicted function values with real function values in Figure \ref{Ex3_PointPrediction}. 
In each subplot, the horizontal-axis gives the parameters that we use to generate input data for the SNN, and the vertical-axis shows function values at the fixed spatial point. We use blue boxes to demonstrate the predicted function values by using our estimated parameters corresponding to the input data generated by the true parameters indicated on the horizontal-axis, and the blue error bars reflect the $95\%$ confidence bands of SNN output in predicting function values. Since the real function values are perturbed by noises with known standard deviation, we use red boxes to represent real function values corresponding to the parameters shown on the horizontal axis, and the red error bars are $95\%$ error bands of function values caused by the perturbation noises. We can see from both subplots that our predicted function values are very close to real function values for all the parameters that we tested, and the $95\%$ confidence bands of predicted function values always cover the real function values. We also notice that the $95\%$ confidence bands of function prediction are wider than the $95\%$ error bands of the real function values. The reason that the SNN produces more uncertainties than the noises in data is that we selected limited random spatial points to calculate data of function values --- both in the training process and in the testing process. The randomness involved in the spatial points selection is incorporated into the SNN, which results wider confidence bands of function prediction.

\section{Conclusion and future work}\label{Conclusion}
In this paper, we introduced a probabilistic machine learning approach, which formulates an SNN model by a stochastic optimal control problem. The numerical method that we developed to train the SNN is an efficient SGD algorithm that implements the SMP approach for optimal control problems. Three numerical examples are provided to validate our methodology in learning performance.
In the future, we plan to carry out rigorous convergence analysis for our computational framework. We also plan to implement our SMP approach for SNN to solve more practical large-scale scientific problems, where probabilistic machine learning is needed.

\bibliographystyle{plain}

\bibliography{Reference}

\vspace{2em}



\begin{thebibliography}{10}

\bibitem{chen2018neural}
Ricky~TQ Chen, Yulia Rubanova, Jesse Bettencourt, and David~K Duvenaud.
\newblock Neural ordinary differential equations.
\newblock In {\em Advances in neural information processing systems}, pages
  6571--6583, 2018.

\bibitem{Hamiltonian-MC}
Tianqi Chen, Emily~B. Fox, and Carlos Guestrin.
\newblock Stochastic gradient hamiltonian monte carlo.
\newblock In {\em Proceedings of the 31st International Conference on Machine
  Learning}, 2014.

\bibitem{ML-Approximation18}
Bo~Dai, Albert Shaw, Lihong Li, Lin Xiao, Niao He, Zhen Liu, Jianshu Chen, and
  Le~Song.
\newblock {SBEED}: Convergent reinforcement learning with nonlinear function
  approximation.
\newblock volume~80 of {\em Proceedings of Machine Learning Research}, pages
  1125--1134, Stockholmsmässan, Stockholm Sweden, 10--15 Jul 2018. PMLR.

\bibitem{BSDE_finance}
N.~El~Karoui, S.~Peng, and M.~C. Quenez.
\newblock Backward stochastic differential equations in finance.
\newblock {\em Math. Finance}, 7(1):1--71, 1997.

\bibitem{fang2019sharp}
Cong Fang, Zhouchen Lin, and Tong Zhang.
\newblock Sharp analysis for nonconvex {SGD} escaping from saddle points.
\newblock In {\em Conference on Learning Theory}, pages 1192--1234, 2019.

\bibitem{Feng_2013}
Xiaobing Feng, Roland Glowinski, and Michael Neilan.
\newblock Recent developments in numerical methods for fully nonlinear second
  order partial differential equations.
\newblock {\em SIAM Rev.}, 55(2):205--267, 2013.

\bibitem{Nature-Learning}
Zoubin Ghahramani.
\newblock Probabilistic machine learning and artificial intelligence.
\newblock {\em Nature}, 521:pages452–459, 2015.

\bibitem{Gong_2017}
Bo~Gong, Wenbin Liu, Tao Tang, Weidong Zhao, and Tao Zhou.
\newblock An efficient gradient projection method for stochastic optimal
  control problems.
\newblock {\em SIAM J. Numer. Anal.}, 55(6):2982--3005, 2017.

\bibitem{Haber_2017}
Eldad Haber and Lars Ruthotto.
\newblock Stable architectures for deep neural networks.
\newblock {\em Inverse Problems}, 34(1):014004, dec 2017.

\bibitem{CNN_Control}
Eldad Haber, Lars Ruthotto, Elliot Holtham, and Seong-Hwan Jun.
\newblock Learning across scales - multiscale methods for convolution neural
  networks.
\newblock {\em arXiv:1703.02009v2}, 2017.

\bibitem{ML_BSDE}
Jiequn Han, Arnulf Jentzen, and Weinan E.
\newblock Solving high-dimensional partial differential equations using deep
  learning.
\newblock {\em Proc. Natl. Acad. Sci. USA}, 115(34):8505--8510, 2018.

\bibitem{Learning-Classification}
Kaiming He, Xiangyu Zhang, Shaoqing Ren, and Jian Sun.
\newblock Deep residual learning for image recognition.
\newblock In {\em Proceedings of the IEEE Conference on Computer Vision and
  Pattern Recognition (CVPR)}, June 2016.

\bibitem{BNN-ICML-2015}
Jose~Miguel Hernandez-Lobato and Ryan~P. Adams.
\newblock Probabilistic backpropagation for scalable learning of bayesian
  neural networks.
\newblock In {\em Proceedings of the 32nd International Conference on Machine
  Learning}, 2015.

\bibitem{jain2017non}
Prateek Jain and Purushottam Kar.
\newblock Non-convex optimization for machine learning.
\newblock {\em Foundations and Trends{\textregistered} in Machine Learning},
  10(3-4):142--336, 2017.

\bibitem{Neural-Jump-SDE}
Junteng Jia and Austin Benson.
\newblock Neural jump stochastic differential equations.
\newblock In {\em 33rd Conference on Neural Information Processing Systems},
  2019.

\bibitem{Learning-approximation}
Patrick Kidger and Terry Lyons.
\newblock {Universal Approximation with Deep Narrow Networks}.
\newblock volume 125 of {\em Proceedings of Machine Learning Research}, pages
  2306--2327. PMLR, 09--12 Jul 2020.

\bibitem{SDE-Net}
Lingkai Kong, Jimeng Sun, and Chao Zhang.
\newblock Sde-net: Equipping deep neural networks with uncertainty estimates.
\newblock In {\em Proceedings of the 37th International Conference on Machine
  Learning}, 2020.

\bibitem{Neural-SDE}
Xuanqing Liu, Tesi Xiao, Si~Si, Qin Cao, Sanjiv~Kumar Kumar, and Cho-Jui Hsieh.
\newblock Neural sde: Stabilizing neural ode networks with stochastic noise.
\newblock In {\em arXiv preprint arXiv:1906.02355}, 2019.

\bibitem{Four_step}
Jin Ma, Philip Protter, and Jiong~Min Yong.
\newblock Solving forward-backward stochastic differential equations
  explicitly---a four step scheme.
\newblock {\em Probab. Theory Related Fields}, 98(3):339--359, 1994.

\bibitem{ma2002representation}
Jin Ma and Jianfeng Zhang.
\newblock Representation theorems for backward stochastic differential
  equations.
\newblock {\em The annals of applied probability}, 12(4):1390--1418, 2002.

\bibitem{Milstein_BSDE}
G.~N. Milstein and M.~V. Tretyakov.
\newblock Numerical algorithms for forward-backward stochastic differential
  equations.
\newblock {\em SIAM J. Sci. Comput.}, 28(2):561--582, 2006.

\bibitem{Morzfeld_Parameter}
Matthias Morzfeld, Marcus~S. Day, Ray~W. Grout, George Shu~Heng Pau, Stefan~A.
  Finsterle, and John~B. Bell.
\newblock Iterative importance sampling algorithms for parameter estimation.
\newblock {\em SIAM J. Sci. Comput.}, 40(2):B329--B352, 2018.

\bibitem{Peng_control}
S.~Peng.
\newblock A general stochastic maximum principle for optimal control problems.
\newblock {\em SIAM J. Control Optim.}, 28(4):966--979, 1990.

\bibitem{Control_survey}
Huy\^{e}n Pham.
\newblock On some recent aspects of stochastic control and their applications.
\newblock {\em Probab. Surv.}, 2:506--549, 2005.

\bibitem{BNN-NIPS2016}
Jost~Tobias Springenberg, Aaron Klein, Stefan Falkner, and Frank Hutter.
\newblock Bayesian optimization with robust bayesian neural networks.
\newblock In D.~D. Lee, M.~Sugiyama, U.~V. Luxburg, I.~Guyon, and R.~Garnett,
  editors, {\em Advances in Neural Information Processing Systems 29}, pages
  4134--4142. Curran Associates, Inc., 2016.

\bibitem{weinan2019mean}
E~Weinan, Jiequn Han, and Qianxiao Li.
\newblock A mean-field optimal control formulation of deep learning.
\newblock {\em Research in the Mathematical Sciences}, 6(1):10, 2019.

\bibitem{Langevin-MC}
Max Welling and Yee~Whye Teh.
\newblock Bayesian learning via stochastic gradient langevin dynamics.
\newblock In {\em Proceedings of the 28th International Conference on Machine
  Learning}, 2011.

\bibitem{Yong-Zhou-1999}
Jiongmin Yong and Xun~Yu Zhou.
\newblock {\em Stochastic controls}, volume~43 of {\em Applications of
  Mathematics (New York)}.
\newblock Springer-Verlag, New York, 1999.
\newblock Hamiltonian systems and HJB equations.

\bibitem{ZhangJ_BSDE}
Jianfeng Zhang.
\newblock A numerical scheme for {BSDE}s.
\newblock {\em Ann. Appl. Probab.}, 14(1):459--488, 2004.

\bibitem{Zhao_BSDE}
Weidong Zhao, Lifeng Chen, and Shige Peng.
\newblock A new kind of accurate numerical method for backward stochastic
  differential equations.
\newblock {\em SIAM J. Sci. Comput.}, 28(4):1563--1581, 2006.

\end{thebibliography}

\end{document}